\documentclass[10pt,twocolumn,letterpaper]{article}

\usepackage[pagenumbers]{cvpr} %

\usepackage{graphicx}
\usepackage{amsmath}
\usepackage{amssymb}
\usepackage{booktabs}
\usepackage{algorithm2e}
\usepackage{amsfonts}
\usepackage{xcolor}
\usepackage[accsupp]{axessibility}  %

\newcommand{\besizer}{b$\mathrm{\acute{e}}$zier }
\SetKwInOut{KwIn}{Input}
\SetKwInOut{KwOut}{Output}
\RestyleAlgo{ruled}

\usepackage[pagebackref,breaklinks,colorlinks]{hyperref}

\usepackage[capitalize]{cleveref}
\crefname{section}{Sec.}{Secs.}
\Crefname{section}{Section}{Sections}
\Crefname{table}{Table}{Tables}
\crefname{table}{Tab.}{Tabs.}

\usepackage[normalem]{ulem}

\def\cvprPaperID{6116} %
\def\confName{CVPR}
\def\confYear{2022}

\begin{document}

\title{Towards Layer-wise Image Vectorization}

\author{
Xu Ma\textsuperscript{1}, Yuqian Zhou\textsuperscript{2,3*}, Xingqian Xu\textsuperscript{2*},
Bin Sun\textsuperscript{1},
\\
Valerii Filev\textsuperscript{4}, Nikita Orlov\textsuperscript{4}, Yun Fu\textsuperscript{1}, Humphrey Shi\textsuperscript{2,4} 
\\
{\small \textsuperscript{1}Northeastern University, \textsuperscript{2}UIUC, \textsuperscript{3}Adobe  Inc. \textsuperscript{4}Picsart AI Research (PAIR)
}
}
\renewcommand{\thefootnote}{\fnsymbol{footnote}}

\twocolumn[{
\renewcommand\twocolumn[1][]{#1}
\maketitle
\begin{center}
    \centering
    \renewcommand{\arraystretch}{0.05} %
    \vspace{-0.35cm}
    \begin{tabular}{c@{\hspace{0.005\linewidth}}c}
       \includegraphics[width = 0.99\linewidth]{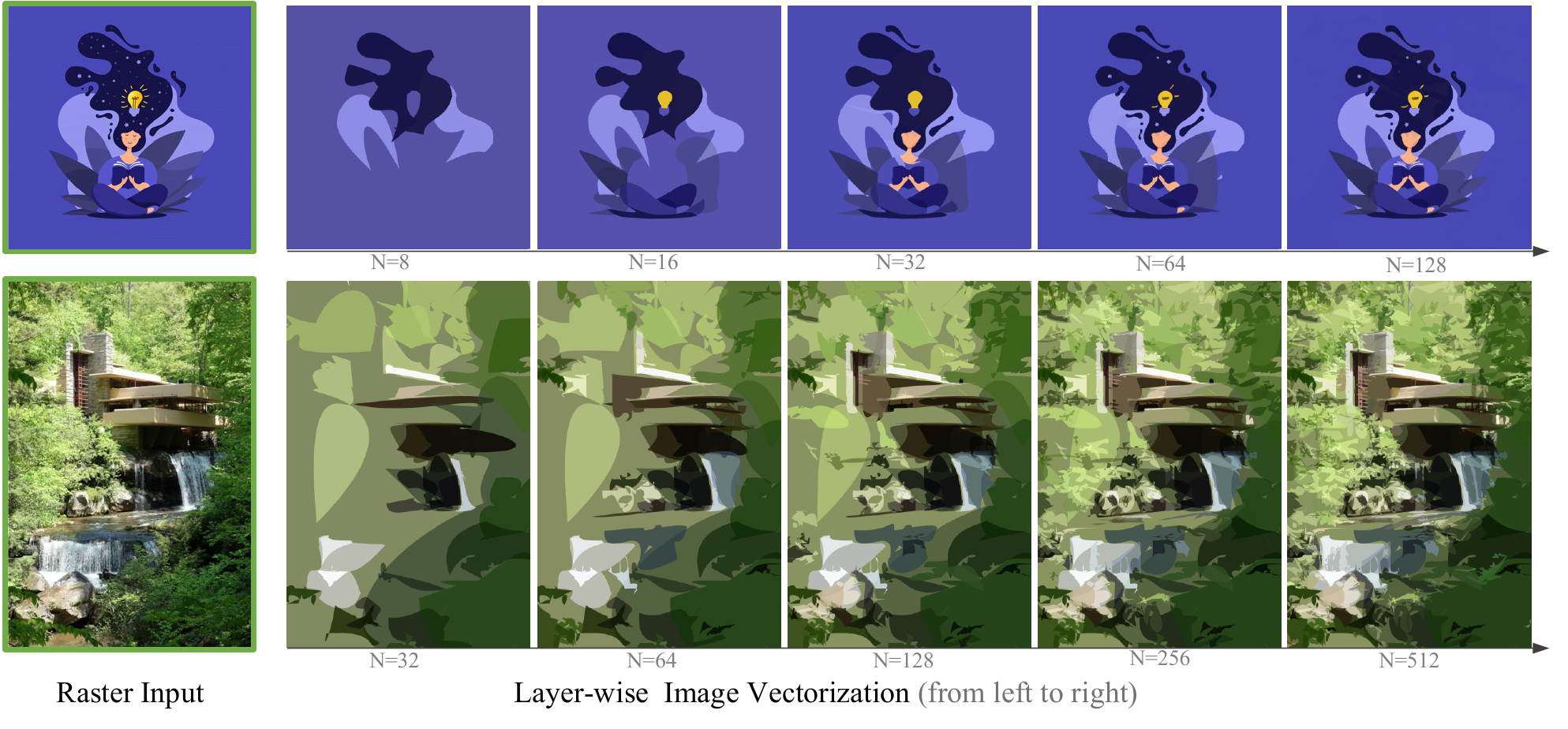}
    \end{tabular}
  \vspace{-0.3cm}
  \captionof{figure}{Examples of the learning course of our Layer-wise Image Vectorization. The proposed method can reconstruct the image in a layer-wise coarse-to-fine manner with only few paths. ``N" indicates the path number. } 
  \label{fig:first}
\end{center}
}]
\footnotetext[1]{Yuqian and Xingqian contributed equally.}

\begin{abstract}
Image rasterization is a mature technique in computer graphics, while image vectorization, the reverse path of rasterization, remains a major challenge. Recent advanced deep learning-based models achieve vectorization and semantic interpolation of vector graphs and demonstrate a better topology of generating new figures. However, deep models cannot be easily generalized to out-of-domain testing data. The generated SVGs also contain complex and redundant shapes that are not quite convenient for further editing. 
Specifically, the crucial layer-wise topology and fundamental semantics in images are still not well understood and thus not fully explored.
In this work, we propose Layer-wise Image Vectorization, namely LIVE, to convert raster images to SVGs and simultaneously maintain its image topology. 
LIVE can generate compact SVG forms with layer-wise structures that are semantically consistent with human perspective. We progressively add new \besizer paths and optimize these paths with the layer-wise framework, newly designed loss functions, and component-wise path initialization technique.
Our experiments demonstrate that LIVE presents more plausible vectorized forms than prior works and can be generalized to new images. 
With the help of this newly learned topology,  LIVE initiates human editable SVGs for both designers and other downstream applications. Codes are made available at \href{https://github.com/Picsart-AI-Research/LIVE-Layerwise-Image-Vectorization}{https://github.com/Picsart-AI-Research/LIVE-Layerwise-Image-Vectorization}.
\end{abstract}

\section{Introduction}\label{sec:intro}
Scalable Vector Graphics (SVGs)~\cite{quint2003scalable}, which describe images with a collection of parametric shape primitives, have recently attracted increasing attention due to the high practical value in computer graphics. 
Compared with raster images that present the visual concepts using ordered pixels, vector images enjoy many advantages, like compact file size and resolution-independency. 
Most importantly, vector images provide layer-wise topological information, which is crucial for image understanding and editing. 

In the last few years, we have witnessed various achievements in image-to-vector translation~\cite{li2020differentiable,carlier2020deepsvg,reddy2021im2vec,egiazarian2020deep,shen2021clipgen,lopes2019learned}, mainly due to advances in two technical directions: building powerful generation models, and employing decent differentiable rendering methods.
These methods, despite their promising vectorization and generation ability, have always overlooked the topological information hidden behind the raster images.
The missing of such information always incurs inadequate learning of vectorization and requires superfluous shape primitives to make up~\cite{li2020differentiable,zou2021stylized,liu2021paint}.
Some methods attempt to resolve this dilemma by either focusing on particular simple datasets~\cite{reddy2021im2vec,reddy2021multi} or employing a segmentation pre-processing method~\cite{favreau2017photo2clipart,egiazarian2020deep}, but each one has its own drawbacks and subtleties. 
The first line of work learns to explore the geometric information of fonts or emojis but cannot be generalized to broad domains.
The other line that considers segmentation pre-processing method requires heavy pre-processing operations and would segment high-contrast texture into multiple small regions, resulting in redundancy~\cite{favreau2017photo2clipart}. 
Hence, a simple yet effective method is desired in the community to capture the layer-wise representation for image-to-vector translation.

In this paper, we introduce a \textbf{L}ayer-wise \textbf{I}mage \textbf{VE}ctorization method, termed as LIVE, to translate a raster image to vector graphics (\textit{i.e.}, SVG) with layer-wise representation.
Different from previous works~\cite{liu2021paint,reddy2021im2vec}, LIVE is model-free and requires no shape primitive labels. 
This property helps us to escape the regime of particular domains like fonts and emojis, and bypass the difficulty of SVG dataset collection or generalization. 
Moreover, LIVE enjoys an intuitive and succinct learning course.
In each step, we are in pursuit of maximizing the topology exploration rather than only minimizing the pixel-wise difference.
The key insight behind this idea is that simply minimizing the vectorization error (\textit{e.g.}, MSE loss between an input raster image and rendered vector graphics) for optimization would lead to a color mean error. %
We achieve this by a component-wise path initialization method and a novel Unsigned Distance guided Focal loss function (UDF loss).
Besides, to mitigate the self-interaction issue, which always occurs in the optimization course ~\cite{reddy2021im2vec}, we present a novel Self-Crossing loss (Xing loss) by adding constraints to the control points optimization.

We evaluate our proposed method for various tasks, including image-to-vector translation and interpolation across domains (\textit{e.g.}, cliparts, emojis, photos, and natural images) to showcase the effectiveness of LIVE.
Our main contributions in this work can be summarized as follows:
\begin{itemize}
    \item We propose LIVE, a general image vectorization pipeline that hierarchically optimizes the vector graph in a layer-wise manner. Our rendering scheme is fully differentiable and can generate layer-wise SVGs which are largely consistent with human perception.
    \item Together with LIVE, we also introduce a general initialization method and novel loss functions, including Self-Crossing loss (Xing loss) and Unsigned Distance guided Focal loss (UDF loss). 
    These methods improve the generation of SVGs from raster images, reducing curve intersection and minimizing the shape distortion.
    \item Comprehensive experiments demonstrate that LIVE can generate precise and compact SVGs in various domains. 
    Our SVG results surpass the results from prior works in terms of simplicity and layer-wise topology.
    
\end{itemize}

\section{Related Work}\label{sec:relatedwork}
In this section, we mainly summarize prior approaches and introduce works that are closely related to our paper. 

\subsection{Rasterization and Vectorization}

Rasterization and vectorization are dual problems in computer graphics. In the past decades, many rasterization works focused on either effective rendering~\cite{ras-effective-01, ras-effective-02, ras-effective-03, ras-effective-04} or anti-aliasing~\cite{ras-antialias-01, ras-antialias-02, ras-antialias-03, ras-antialias-04}. Traditional vectorization methods~\cite{vec-seg-01, vec-seg-02, vec-seg-03, vec-seg-04, vec-seg-05, vec-seg-06} pre-segmented images before vectorization. Among them, \cite{vec-seg-01} and \cite{vec-seg-05} utilized the empirical two-stage algorithm to regress segmented components as polygons and bezigons. Researchers also investigated other approaches that were independent of segmentation, such as diffusion curves~\cite{vec-diffuse-01, vec-diffuse-02, vec-diffuse-03} and gradient meshes~\cite{vec-gmesh}. The rise of deep learning motivated researchers to tackle vectorization via differentiable rendering. Yang \etal~\cite{yang2015effective} proposed that bezigons can be directly optimized with self-crafted loss functions by computing gradients using wavelet rasterization~\cite{ras-antialias-04}. Li \etal~\cite{li2020differentiable} found shape gradients by differentiating the formula of Reynolds transport theorem~\cite{rtt} with Monta-Carlo edge sampling. 
Meanwhile, combining differentiable rendering techniques with deep learning models is a trending research direction. New networks that based on recurrent neural network~\cite{vec-rnn-01}, variational autoencoder~\cite{vec-rnn-01, vec-vae-02}, and transformer~\cite{vec-transformer-03} are introduced to tackle vectorization and vector graph generation problems. In~\cite{vec-rnn-01}, Ha \etal introduce SketchRNN, which is the first RNN-based sketch generation network. In~\cite{vec-vae-02}, Lopes \etal introduce an SVG decoder and combine it with a pixel VAE to generate novel font SVGs in latent space. 
In~\cite{vec-transformer-03}, Rebeiro \etal propose Sketchformer, a transformer-based network that recovers sketches from raster images. 

\subsection{Image Topology}

A human editable SVG should be well-organized in objects and shapes. Prior works have explored such image topology for both raster images and vectorized shapes. A prototype work was Photo2ClipArt~\cite{favreau2017photo2clipart}, where images were first split into segments which were later vectorized then combined for visual hierarchy. Similar design repeatedly occurs at other research works such as~\cite{shimoda2021rendering,shen2021clipgen}.
Nevertheless, these methods were largely relied on the accuracy of the segmentation step and were clumsy to recover implicit shape geometry for complex scenes.
Another research branch designed end-to-end frameworks to generate or edit image hierarchy through one forward pass. For example, DeepSVG~\cite{carlier2020deepsvg} used VAE as its primary structure where input strokes are first represented via an encoder and later replaced by resampled strokes via a decoder. However, DeepSVG performed SVG to SVG translation, a relatively simple task. Stylized Neural Painting~\cite{zou2021stylized} reconstructed images progressively with stylized strokes. Their design principle was to greedy-search for the best stroke to minimize the loss. Yet their main focus was on raster images, not SVG. Lastly, Im2Vec~\cite{reddy2021im2vec} proposed the Encoder-RNN-Rasterizer pipeline to vectorize an image and obtain its topology simultaneously. However, the ordering of the generated shapes was not robust, and the method was domain-specific. Different from aforementioned methods, our LIVE requires no pre-segmentation and no deep models, but exhibits gratifying ability to explore image topologies.

\section{LIVE: Layer-wise Image Vectorization}\label{sec:method}
\subsection{Framework}
We present a new method to progressively generate an SVG that fits the raster image in a layer-wise fashion. Given an arbitrary input image, LIVE recursively learns the visual concepts by adding new optimizable closed \besizer paths and optimizing all these paths.
While various shape primitives are available to be appended to an SVG, we consider the parametric closed \besizer path as our fundamental shape primitive, like the implementation in~\cite{reddy2021im2vec,li2020differentiable}. 
There are several reasons behind this setting. 
First, this strategy would greatly reduce the design space and significantly ease the learning course of LIVE.
Also, \besizer paths are powerful and easy to approximate diverse shapes, making it unnecessary to introduce various shape primitives.
Last, it is convenient for us to control the shape complexity by varying the number of segments $s$ in each path. 
For complex visual concepts, we can easily increase the segment number to better reconstruct input, and vice versa.
Note that the rendering operation is usually non-differentiable, making it difficult to directly optimize the path under the only supervision of the target raster image.
To grapple with this dilemma, we take the advantage of a differentiable renderer from~\cite{li2020differentiable}.

Algorithm.~\ref{alg:algorithm} shows the entire pipeline.
Briefly, we first introduce a component-wise initialization method that selects the major components as the initialization points. Then we run a recursive pipeline to progressively add $n$ paths according to a path number scheduler sequence $\mathbb{N}$. For each step, we optimize the graph based on some newly proposed objective functions, including an Unsigned Distance guided Focal (UDF) loss and a Self-Crossing (Xing) loss for a better optimization result regarding the reconstruction quality and self-interaction problem.
In addition to the layer-wise representation ability, our method is able to reconstruct an image using minimal number of \besizer paths, significantly reducing the SVG file size compared to other methods. More details are covered in the following sections.

\begin{algorithm}[!t]
\caption{Algorithm of LIVE}
\label{alg:algorithm}

P = [] \textcolor{gray}{\tcp*{list of path control points}}
C = [] \textcolor{gray}{\tcp*{list of path colors}}
$w$ =1.0 \textcolor{gray}{\tcp*{pixel-wise loss weight}}
$\alpha,\ \beta = 1.0, \  0.01$\textcolor{gray}{\tcp*{learning rate}}
\For{ n in $\mathbb{N}$ }{
    \textcolor{gray}{\tcp{new points $p\in\mathbb{R}^{n\times 4s\times 2}$ }}
    \textcolor{gray}{\tcp{new colors $c\in{\mathbb{R}^{n\times 4}}$}}
    p, c = $\mathrm{init}\left(n, w\right)$ \;
    P = concat([P;\quad p])\;
    C = concat([C;\quad  c])\;

    \For{$j=1$ to t}{
        $\hat{I} = \mathrm{render}\left(\mathrm{P},\mathrm{C}\right)$\;
        $\mathcal{L} = \mathcal{L}_{\text{UDF}}\left(I-\hat{I} \right) + \lambda\mathcal{L}_{\text{Xing}}\left(P\right)$\;
        $\mathrm{P} = P-\alpha \frac{d\mathcal{L}\left(\mathrm{P}  \right)}{d\mathrm{P}}$ \textcolor{gray}{\tcp*{update points}}
        $\mathrm{C} = C-\beta \frac{d\mathcal{L}\left(\mathrm{C}  \right)}{d\mathrm{C}}$ \textcolor{gray}{\tcp*{update colors}}
    }
    $w = \left\|I-\hat{I} \right\|_2$  \textcolor{gray}{\tcp*{update $w$}}
}
\KwOut{Scalable Vector Graphic SVG$\left\{ \mathrm{P}, \mathrm{C}\right\}$.}
\end{algorithm}

\subsection{Component-wise Path Initialization}

We find the initialization of \besizer path is crucial in LIVE. A bad initialization will lead to unsuccessful topological extraction and generate redundant shapes. To overcome this defect, we introduce the component-wise path initialization, which greatly helps the optimization course. 

The design principle of the component-wise path initialization is to identify the most suitable initial location of the path based on the color and size of each component. 
One component is one connected area that has a uniform-filled color. 
As we mentioned earlier, LIVE is a progressive learning pipeline. 
Given the SVG output from previous stages, we prioritize our next learning target so that the component is both large and missing. 
We justify such component via the following steps: 
a) We compute the $l_1$ pixel-wise color difference between the current rendered SVG and the ground truth image. 
b) We reject color differences that are smaller than a preset threshold $c_{\alpha}$. Empirically, $c_{\alpha} = 0.1$ in our paper. Pixel regions with color differences smaller than $c_{\alpha}$ are considered to be correctly rendered. 
c) For other regions, we equally quantify all valid color difference values larger than $c_{\alpha}$ into 200 bins. The quantization is approximately uniformly distributed. 
d) Finally, we identify the largest connected component based on the quantization, and we then use its center of mass as our next path initial location. If we want to add $K$ more paths, then we choose the top-$K$ components for next-stage initialization. Note that for each path, we consider the circle initialization method that all control points are initialized uniformly on a circle~\cite{reddy2021im2vec}. Empirically, this simple strategy helps to ease the optimization course and is proved to be helpful.

The merit of our component-wise path initialization is that it maintains a good balance between the color and size of the missing region. Unlike DiffVG~\cite{li2020differentiable} and Neural Painting~\cite{zou2021stylized}, in which the former randomly initializes paths and the later initializes strokes based on MSE, our approach focuses on semantic-influencing components that are independent from its RGB value. While adding new paths to the existing figures, our initialization methods can always identify the largest missing components with similar color, and fill in the major regions.

\subsection{Loss Function}

\subsubsection{UDF Loss for Reconstruction}
\begin{figure}
    \centering
    \includegraphics[width=0.98\linewidth]{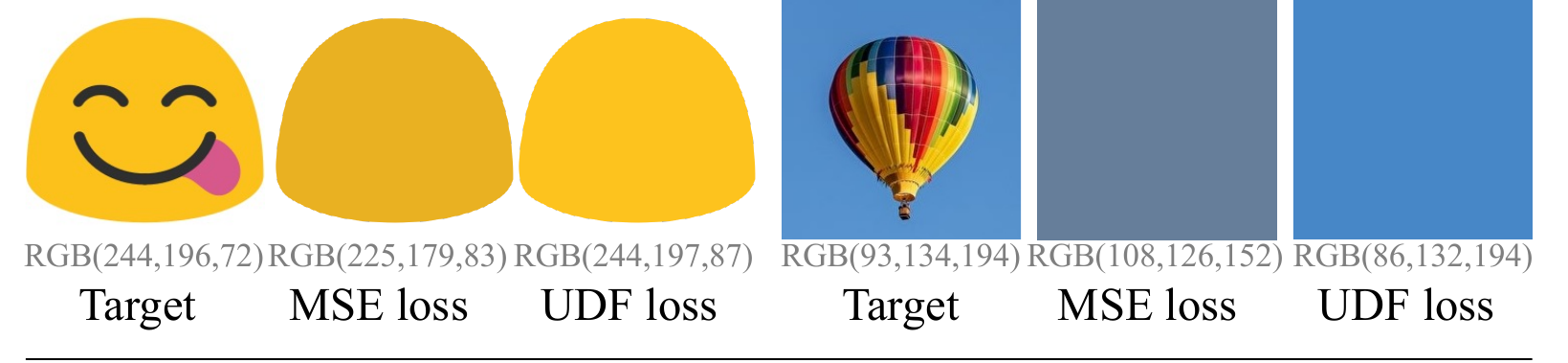}\\
    \vspace{0.3cm}
    \includegraphics[width=0.98\linewidth]{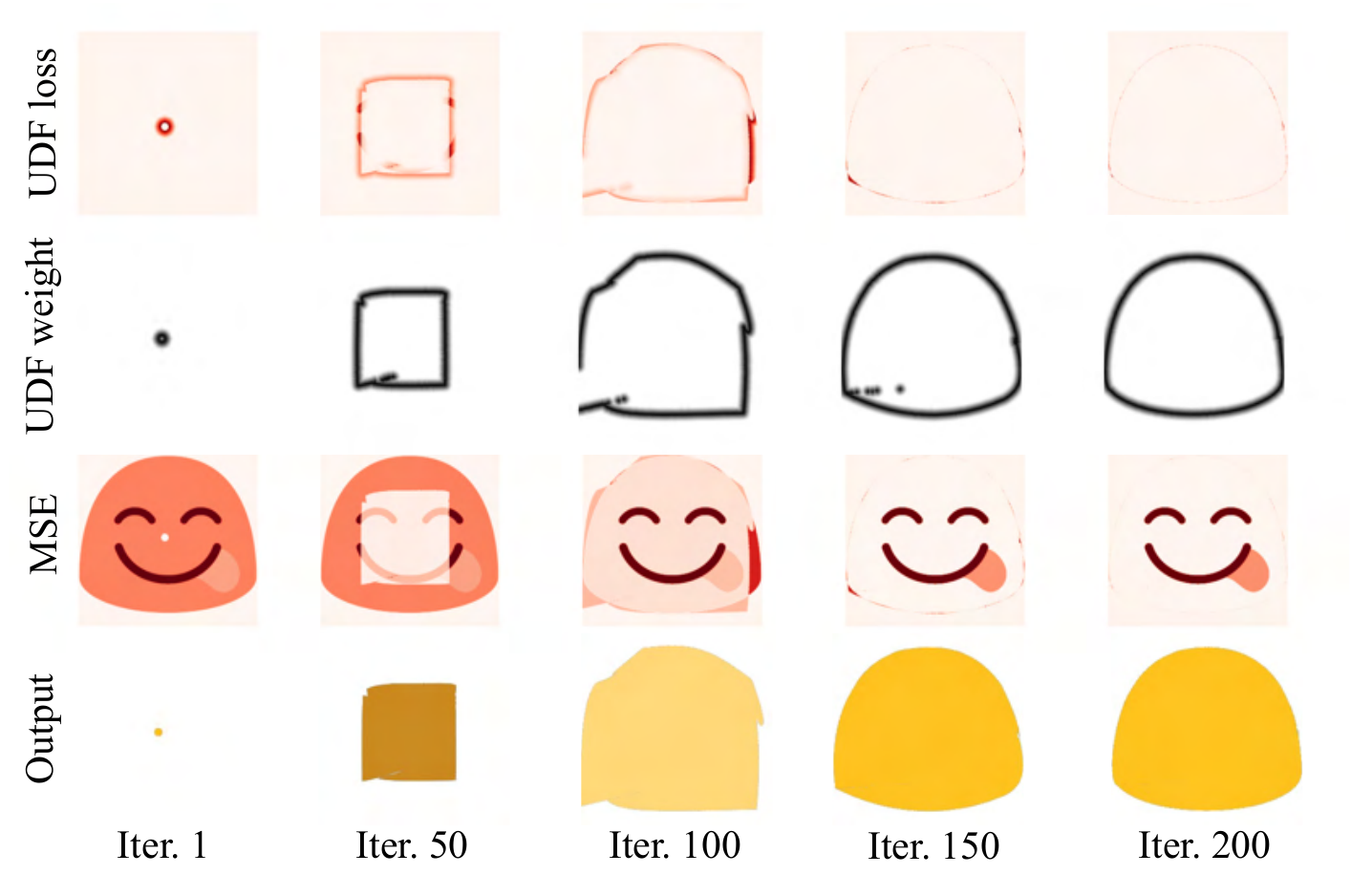}
    \caption{
    \textbf{Top line} shows the differences between UDF loss and MSE loss when learning the first path. MSE loss is biased towards the mean color of the target image, while our UDF loss preserves the color of the target shape. Best viewed in color.
    \textbf{Bottom block} presents an example of UDF loss optimization course for the first path. We normalize all values to the range of $\left[0,1\right]$ for better visualization. Darker color (gray or red) means a higher value.}
    \label{fig:udloss_example}
\end{figure}

In previous work~\cite{li2020differentiable, reddy2021im2vec, reddy2021multi}, a commonly used loss function to minimize the error between target image $I\in\mathbb{R}^{ w\times h \times 3}$ and rendered output $\hat{I}\in\mathbb{R}^{w\times h \times 3}$ is the mean square error (MSE)  $\left \| I-\hat{I} \right \|_2^2$, where $3$ represents RGB  and $w\times h$ represents image size. 
MSE loss is simple yet efficient for image comparison, but it will bias towards the mean color of the entire target image, as shown in Figure~\ref{fig:udloss_example}.
This phenomenon is because MSE are computed using all available pixels, while not all pixels are related to  the optimizing path. Hence, we are encouraged to only focus on  valid pixels and ignore the  unrelated ones.

To  resolve this problem, we introduce the Unsigned Distance guided Focal (UDF) loss, which  treats each pixel differently based on the distance to the shape contour. Intuitively speaking, UDF loss emphasizes the differences close to the contour and suppresses differences in other locations. By doing so, LIVE protects itself from MSE's mean color issue and therefore maintains accuracy color reconstruction.

Without losing generality, we formulate our UDF loss assuming the case with a single path.
We render the path and compute each pixel's signed distance to the path represented by $d_{i}, i\in \{1, ..., h\times w\}$. We then threshold, flip, and normalize the unsigned distance $\left| d_{i} \right|$ by:
\begin{equation}
    {d}'_i= \frac{\mathrm{ReLU}(\tau -\left | d_i \right |))}{\sum_{j=1}^{w\times h} \mathrm{ReLU}(\tau -\left | d_j \right |))},
    \label{eq:loss_weight}
\end{equation}
where both $i$ and $j$ are indices of pixels and $\tau$ is a distance threshold. We set $\tau$ equals to 10 by default. 
Next, we formulate our Unsigned Distance guided Focal loss as

\begin{equation}
   \mathcal{L}_{\text{UDF}} = \frac{1}{3}\sum_{i=1}^{w\times h}{d_i}'\sum_{c=1}^{3}\left ( I_{i,c}-\hat{I}_{i,c} \right )^2,
   \label{eq:sdf_loss}
\end{equation}
where $i$ indexes the pixel in $I$ and $c$ indexes the RGB channel.
With the help of  UDF loss, we are able to pay close attention to the path contour and to avoid the effect from inner or distant regions. 
Figure~\ref{fig:udloss_example} shows the learning course of Unsigned Distance guided Focal loss.
To support multiple paths in our LIVE framework, we can easily extend Equation~\ref{eq:sdf_loss} by averaging $d_i'$ over all paths.

\subsubsection{Xing Loss for Self-Interaction Problem}%
We notice that it is possible for some \besizer paths to become self-interacted during the course of optimization, leading to detrimental artifacts and improper topology~\cite{yang2015effective,reddy2021im2vec}. While it might be expected that additional paths can cover the artifacts, we emphasize this would complicate the generated SVG and cannot effectively explore the underlying topological information. To this end, we introduce the self-interaction (Xing) loss to mitigate this problem.

\begin{figure}
    \centering
    \includegraphics[width=0.98\linewidth]{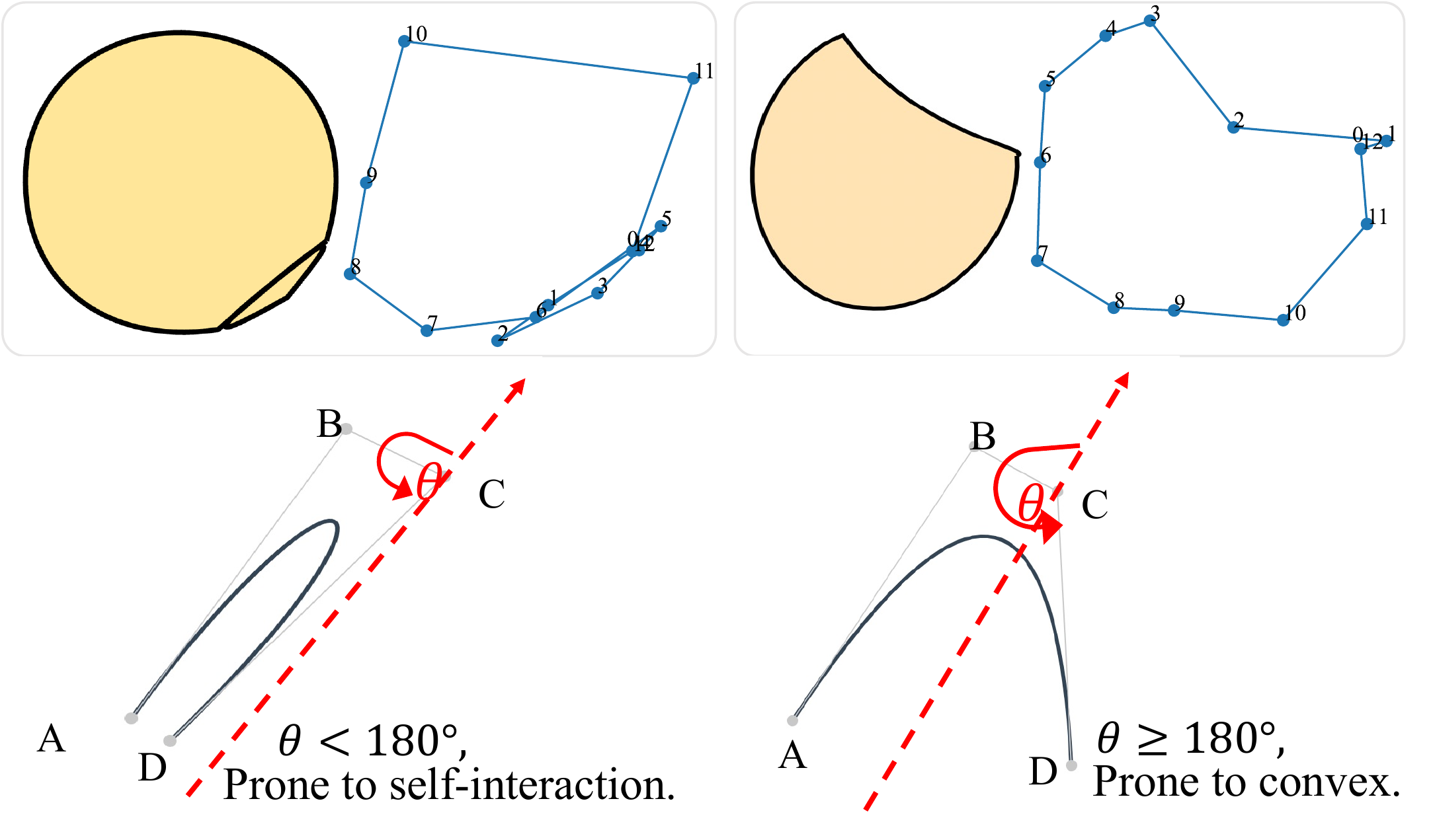}
    \caption{Illustration of self-interaction problem. 
    \textbf{Top left pair} shows a circle with self-interactions and the lines between its adjacent control points.
    \textbf{Top right pair} shows a shape without self-interaction. %
    \textbf{Bottom line} illustrates our Xing loss. 
    In a cubic \besizer path, we encourage the angle ($\theta$) 
    between the first ($\vec{AB} $) 
    and the last ($\vec{CD} $) 
    control points connections greater than $180^{\circ}$.}
    \label{fig:self_interaction_problem}
\end{figure}

Assuming all the \besizer curves in our paper are third-order, by analyzing a number of optimized shapes, we found that a self-interacted path always intersects the lines of its control points, and vice versa. Figure~\ref{fig:self_interaction_problem} shows the examples.
This suggests that instead of optimizing the \besizer path, one potential solution would be adding a constraint on the control points. 
Assume the control points of a cubic \besizer path are $A, B, C,$ and $D$ in sequence, we add a constraint that the angle between $\overrightarrow{AB}$ and $\overrightarrow{CD}$ ($\theta$ in the figure) should be greater than $180^{\circ}$. We first determine the characteristic (acute angle or obtuse angle) of $\angle ABC$ as $D_1$ and the value of $\mathrm{sin}\left(\theta\right)$ as $D_2$ by
\begin{equation}
    D_1 = \mathbb{I}\left(\vec{AB}\times\vec{BC}\right),\quad D_2 = \frac{\vec{AB}\times\vec{CD}}{\left \| \vec{AB} \right \|\left \| \vec{CD} \right \|},
\end{equation}
where $\mathbb{I}\left(\cdot\right)$ is a sign function that returns 1 (if $D_1>0$ ) or 0 (if $D_1\leq0$), $\times$ is vector production that returns a real value.
We then formulate our Xing loss as
{\small
\begin{equation}
    \mathcal{L}_{\text{Xing}} = D_1\left(\mathrm{ReLU}\left(-D_2\right)\right) + \left(1-D_1\right)\left(\mathrm{ReLU}\left(D_2\right)\right).
    \label{eq:xing_loss}
\end{equation}
}

The basic idea of Equation~\ref{eq:xing_loss} is that we only optimize the case when $\theta<180^{\circ}$ (achieved by $\mathrm{ReLU}\left(\pm D_2\right)$). The first term is designed for case $D_1=1$ and the second term is designed for case $D_1=0$.
Combining both UDF loss and Xing loss, our final loss function $\mathcal{L}$ is given by
\begin{equation}
    \mathcal{L} = \mathcal{L}_{\text{UDF}} + \lambda\mathcal{L}_{\text{Xing}},
\end{equation}
where $\lambda$ is set to 0.01 empirically to balance the two losses.

\begin{figure}
    \centering
    \includegraphics[width=0.98\linewidth]{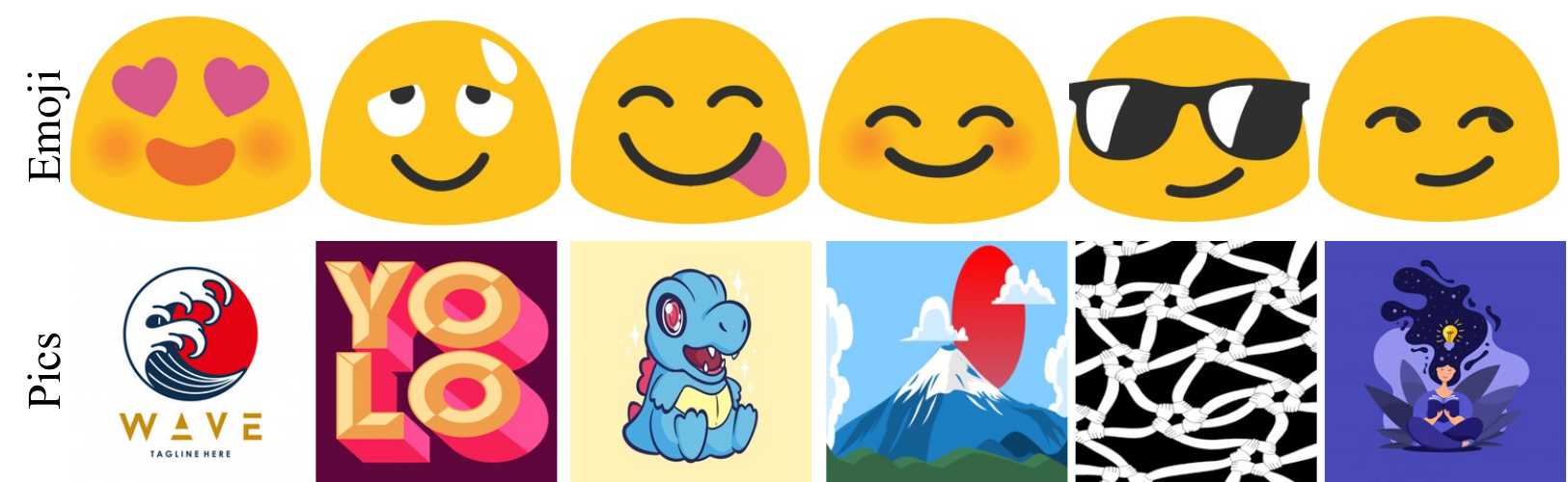}
    \caption{Exemplars from Emoji dataset and Pics dataset.}
    \label{fig:datasets}
    \vspace{-4mm}
\end{figure}

\subsection{Datasets}
Existing vector graphics datasets~\cite{lopes2019learned,carlier2020deepsvg} mainly focus on the generation of either fonts or icons, but a broader domain of images is not explored.  
Also, there is no testing set that can serve as a benchmark for evaluation.
In this paper, we test our model on two datasets, an \textbf{Emoji} dataset that mainly collects a subset of emojis from \cite{Noteemoji}, and a
\textbf{Pics} dataset that collects images from different domains.
Figure~\ref{fig:datasets} showcases some examples from Emoji and Pics datasets.

\paragraph{Emoji Dataset.} We collect 134 emojis with various shapes, colors, and combinations from the NotoEmoji project~\cite{Noteemoji}.
While various fonts and icons are given in this project, we mainly collect the smiling face images, and resize all collected images to a resolution of $240\times240$. 
Compared with the emojis used in~\cite{reddy2021im2vec}, our Emoji dataset includes more images and presents more diversity.
Since images in~\cite{Noteemoji} are relatively simple and present clear topological information, we mainly use this dataset to evaluate the exploration of layer-wise representation.

\paragraph{Pics Dataset}
Besides the Emoji dataset, we also introduce Pics dataset, which contains 153 images, including fonts, icons, and complex clipart images. 
Compared with the Emoji dataset, the Pics dataset is more complex and challenging for image vectorization. Moreover, some images in the Pics dataset are with various backgrounds, further increasing the vectorization difficulty. We mainly use this dataset to examine the layer-wise modeling and compact SVG with fewer paths. 

Note that our LIVE is a model-free method, both datasets are only used to evaluation. Besides the two datasets, we also evaluate LIVE on some realistic photos.

\subsection{Implementation Details}
We implement LIVE in PyTorch~\cite{paszke2019pytorch} and optimize it using Adam optimizer~\cite{kingma2015adam}, with a learning rate of 1 and 0.01 for points and colors optimization, respectively. By default, we use four segments for each path in our experiments. The circle radius is set to 5 pixels for circle initialization. For each optimization step, all the parameters are trained for 500 iterations.  
Since our method progressively adds new paths to the canvas, the number of new paths in each step is flexible. Considering both the efficiency and vectorization quality, we set the path number in $i$th optimization step to be $\mathrm{min}\left ( 2^{i-1}, 32 \right )$. Other number setting strategies also work, like adding one path each time or a customized setting.

\section{Experiments}\label{sec:experiments}

\begin{figure}
    \centering
    \includegraphics[width=0.98\linewidth]{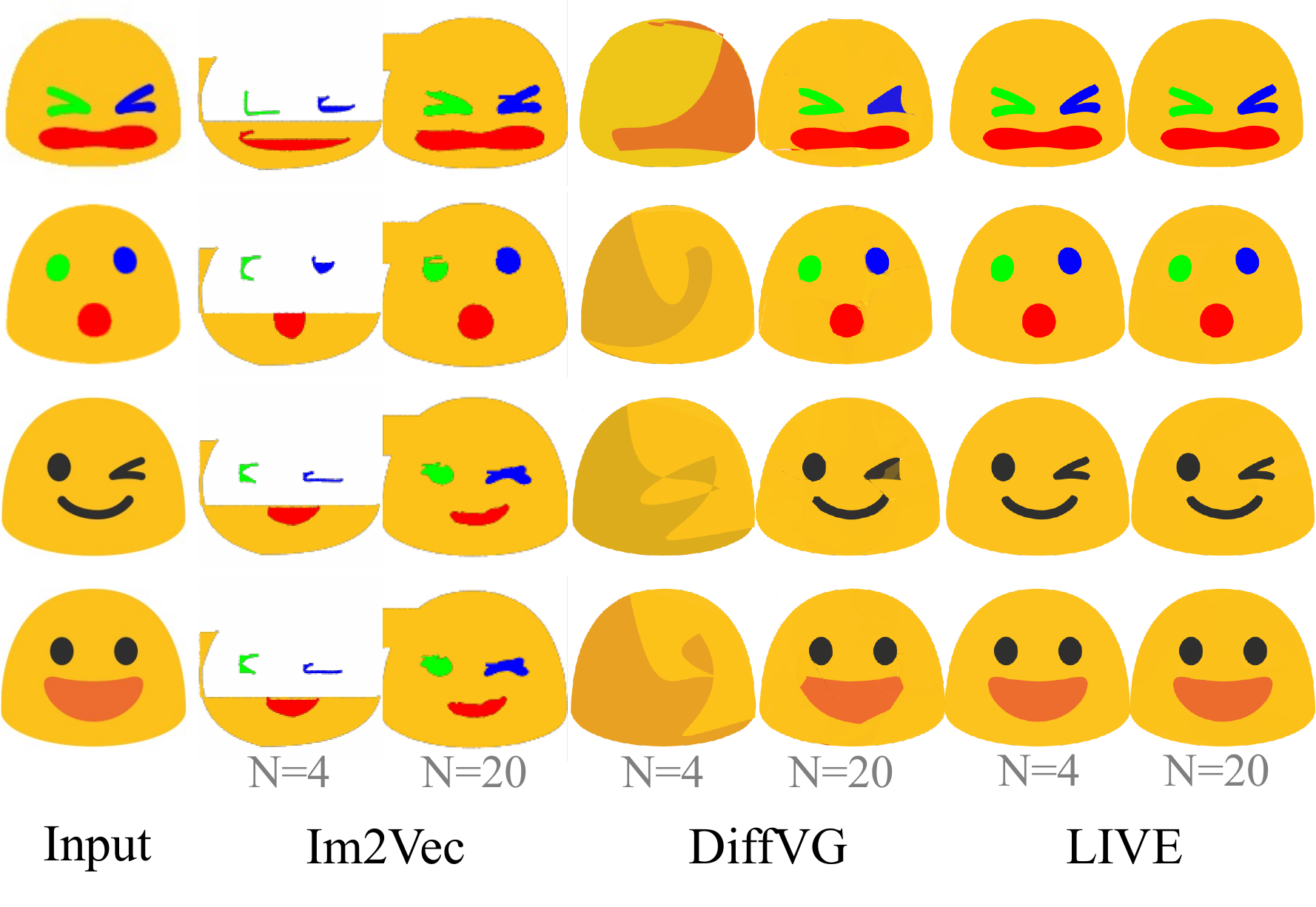}
    \caption{Qualitative reconstruction comparison.
    We compare  LIVE with Im2Vec and DiffVG using a different number of paths.
    We select four paths (the number of components in each image) and 20 paths (default value in Im2Vec) for comparison. Intuitively, LIVE can achieve perfect results using only four paths, and more paths would not degrade the performance.
    }
    \label{fig:compare_no678}
\end{figure}

\subsection{Vectorization Quality}
We first evaluate LIVE's vectorization quality with both quantitative and qualitative analysis, measuring the differences between input targets and SVG rendered images.

\paragraph{Qualitative Comparisons.}
Figure~\ref{fig:compare_no678} shows the visual comparisons with previous state-of-the-art methods including DiffVG \cite{li2020differentiable} and Im2Vec \cite{reddy2021im2vec}. For fairness, we set the number of path to 4 (the number of components in these emojis) and 20 (the default setting in Im2Vec) for evaluation. Clearly, our LIVE achieve a more faithful reconstruction with better component shapes and colors, while others may still have other artifacts. Therefore, the proposed LIVE better decouples the geometry of different components. More results are in the supplementary materials. 

\paragraph{Quantitative Results.}
Next, we quantize the vectorization results on the Emoji and Pics datasets. For a fair comparison, the number of segments is set to 4 as the default setting in DiffVG. To showcase that LIVE can reconstruct one image with a minimal number of paths, we vary the path number from 8 to 64 for the simple Emoji dataset and from 32 to 256 for the complex Pics dataset. For comparison, we calculate the MSE of each image for the entire dataset.

\begin{figure}[!t]
    \centering
    \includegraphics[width=0.49\linewidth]{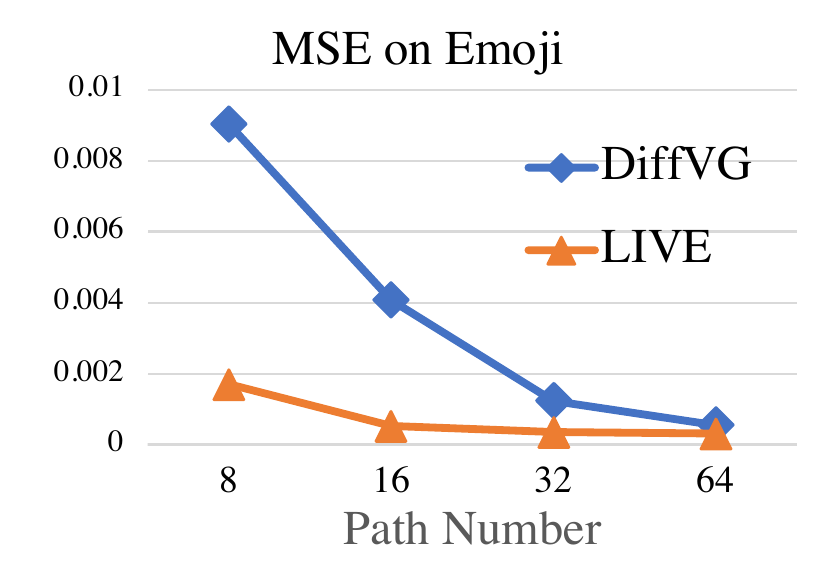}
    \includegraphics[width=0.49\linewidth]{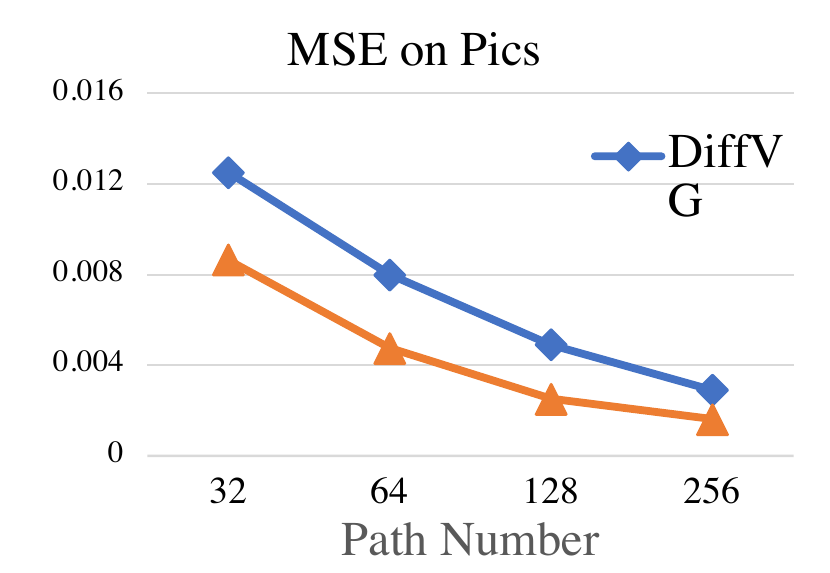}
    \caption{MSE \textit{vs}. path number on Emoji dataset and Pics dataset. Our LIVE achieves much better reconstruction results than DiffVG, especially when the path number is smaller.}
    \label{fig:mse_two_datasets}
\end{figure}

The results on Emoji and Pics benchmarks are reported in Figure~\ref{fig:mse_two_datasets}. Clearly, LIVE shows a much lower MSE than DiffVG, especially when the path number is small. When only a few paths are employed, LIVE is able to fit the desired shapes, leading to a better result. Increasing excessive paths would saturate the vectorization performance.

\begin{figure}[!t]
    \centering
    \includegraphics[width=0.7\linewidth]{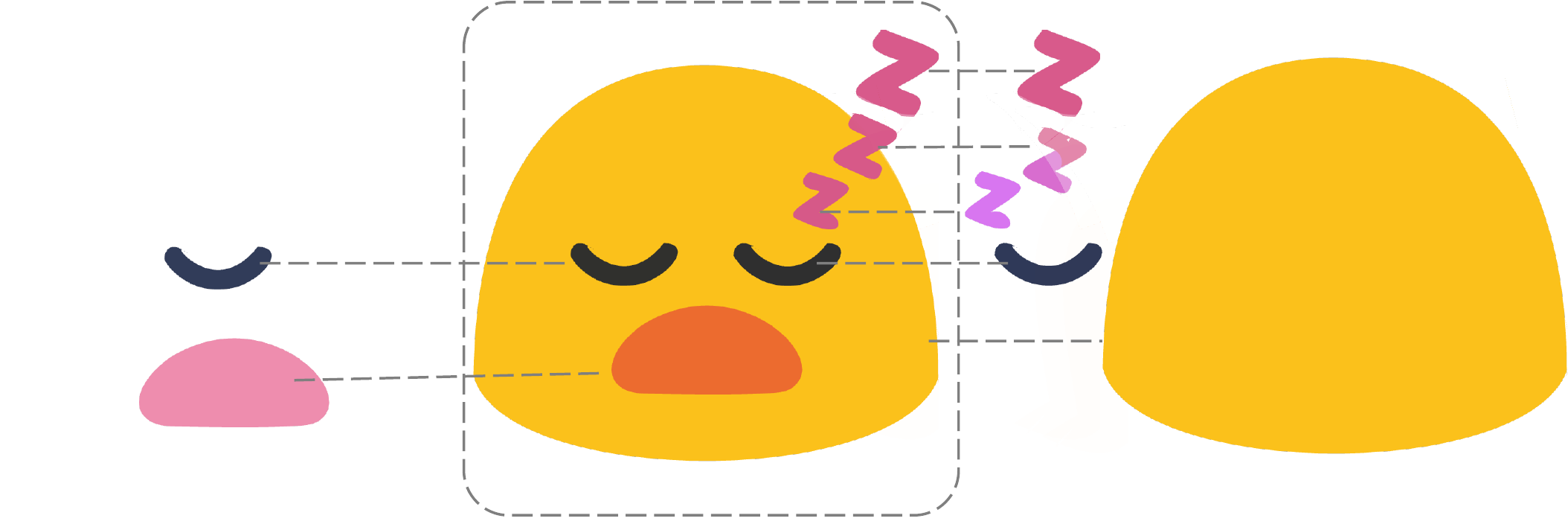}\\
    \includegraphics[width=0.7\linewidth]{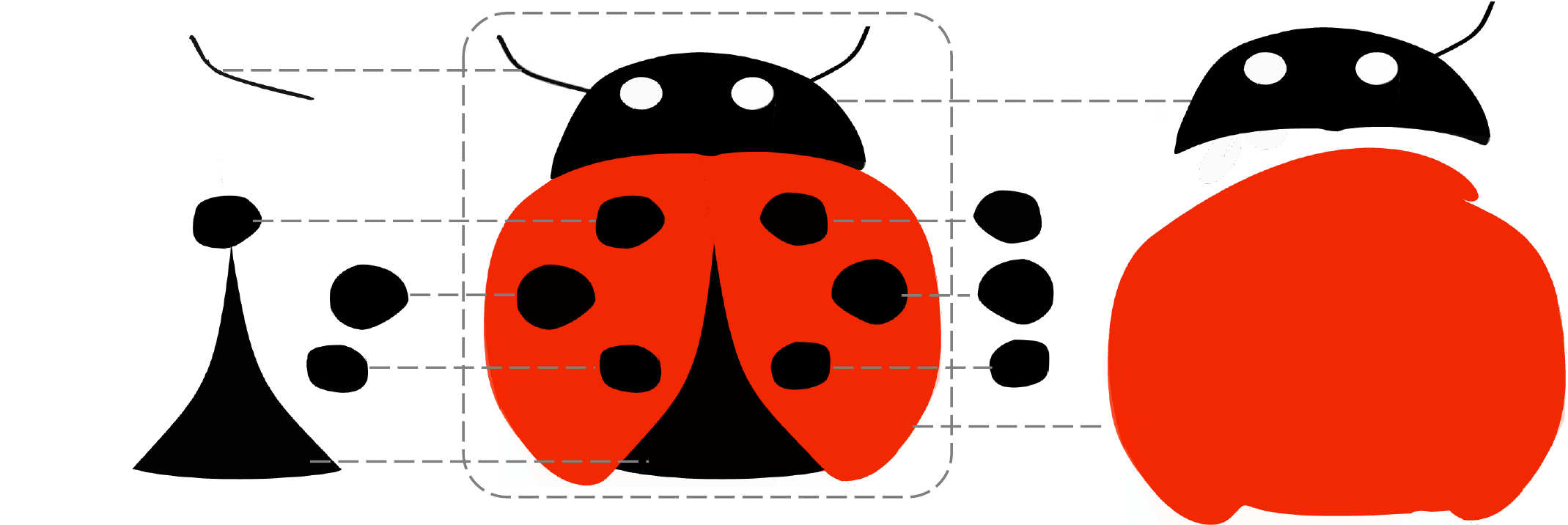}
    \caption{Illustration of the layer-wise representation on emoji and clipart images. When the visual clues are easy to model, LIVE can directly model each individual component, presenting a reasonable and clear layer-wise representation. }
    \label{fig:layerwise_easy}
    \vspace{-4mm}
\end{figure}

\begin{figure*}
    \centering
    
    \includegraphics[width=0.99\linewidth]{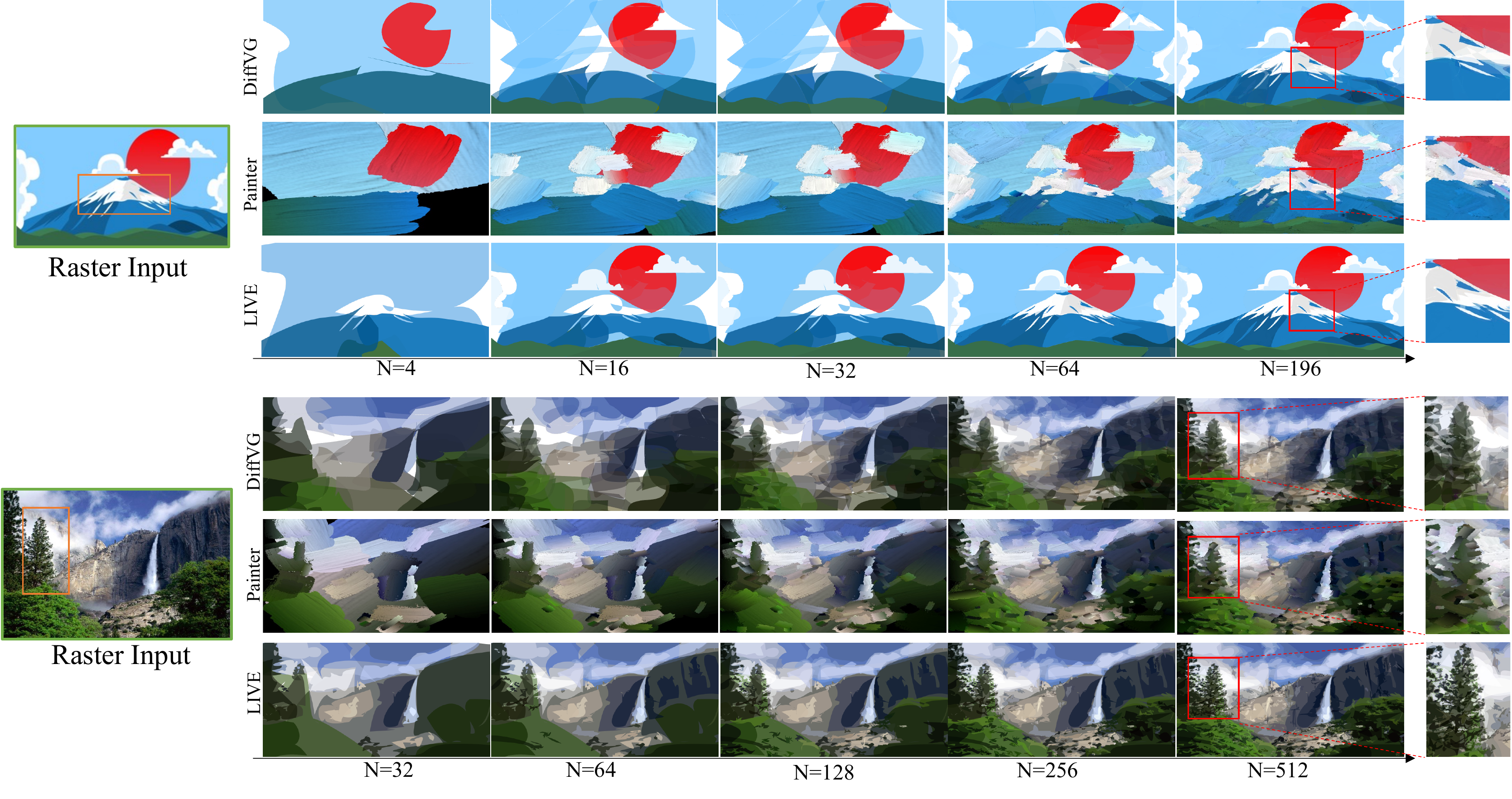} 
    \caption{We showcase the results of DiffVG~\cite{li2020differentiable},  Neural Painting~\cite{zou2021stylized} and our LIVE under different number of paths/strokes. The two images are taken from the Pics dataset and testing images from~\cite{zou2021stylized}, respectively. \textit{Note that Neural Painting is not only designed for reconstruction. We still visually compare with it because of its progressive learning fashion, which is similar to our LIVE. We use red boxes to emphasize the differences.} Please zoom in to see the details. More results will be presented in the supplementary materials.}
    \label{fig:compare_diff_paths}
    \vspace{-4mm}
\end{figure*}

\begin{figure}
    \centering
    \includegraphics[width=0.95\linewidth]{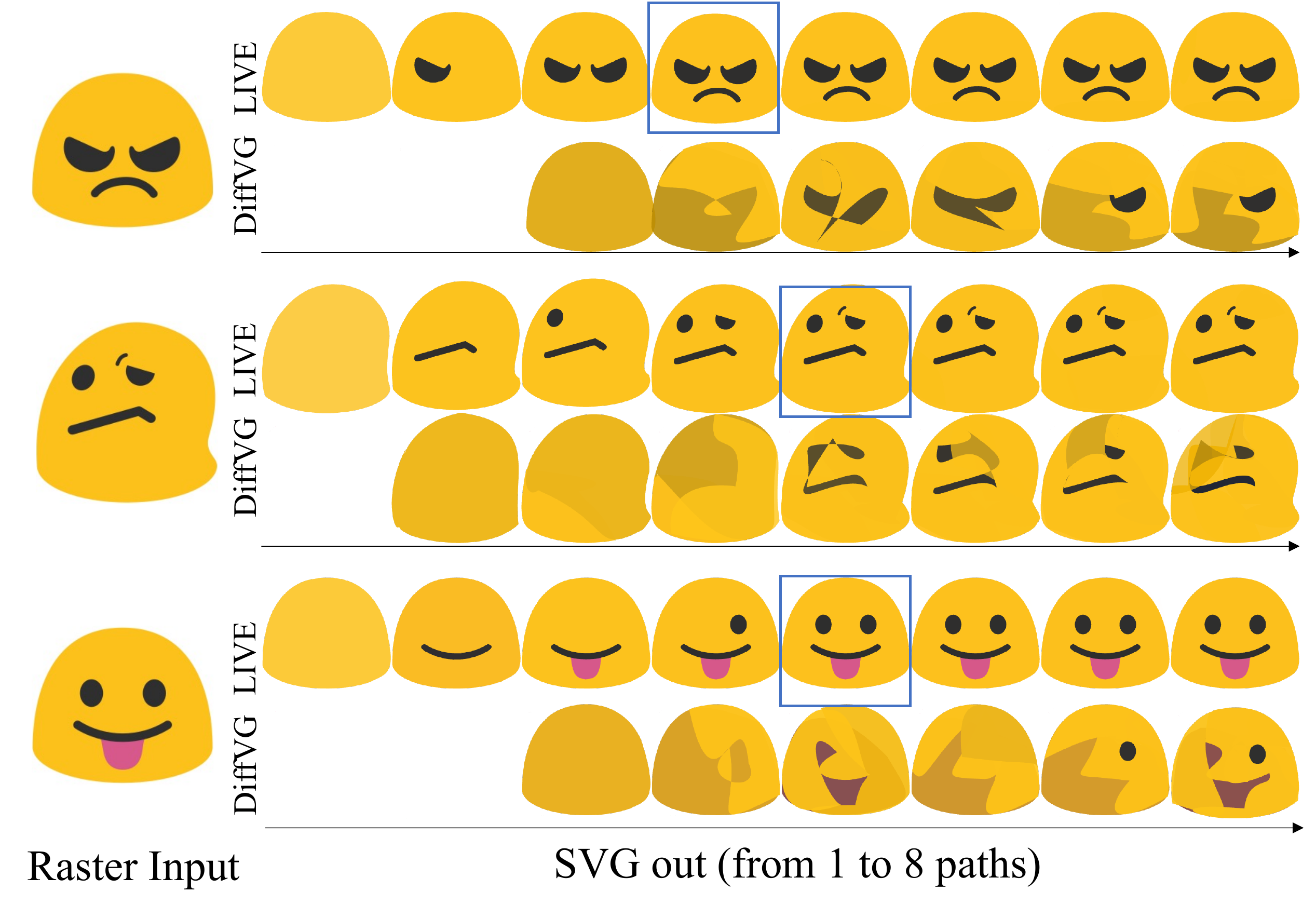}
    \caption{Vectorization results of DiffVG and LIVE. LIVE explicitly vectorized each visual concept, without any redundancy and artifacts. Blue boxes indicate when LIVE vectorized all concepts, and adding more paths will not damage the results. 
    }
    \label{fig:4emoji}
    \vspace{-0.3cm}
\end{figure}

\begin{figure}
    \centering
    \includegraphics[width=0.96\linewidth]{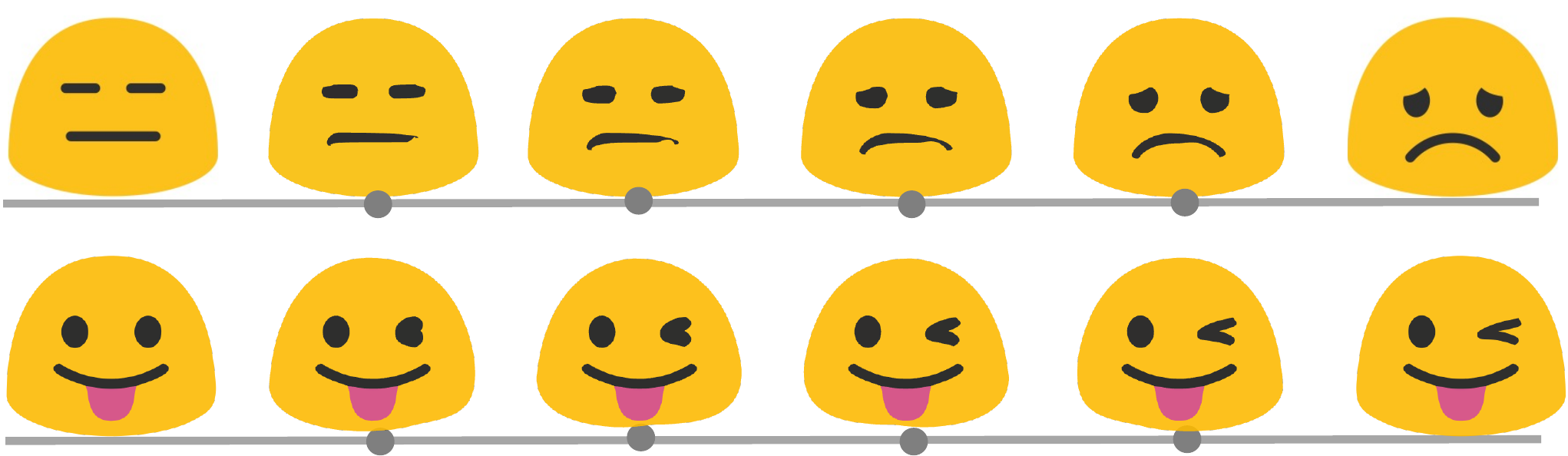}\\
    \vspace{0.4cm}
    \includegraphics[width=0.95\linewidth]{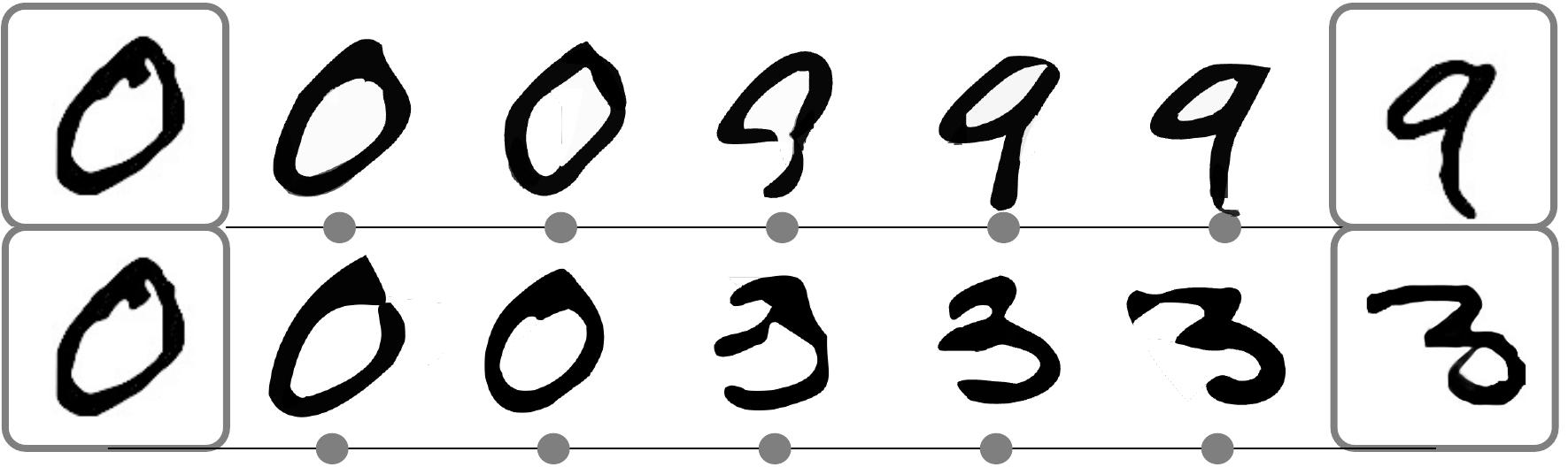}
    \caption{Two examples of interpolation. \textbf{Top tow rows} showcase the results of linear interpolating the \besizer control points between two generated SVGs. \textbf{Bottom two rows} show the results of combining LIVE and a simple VAE. Gray boxes mark the input raster images. Intermediate images indicate the interpolations.}
    \label{fig:interpolation}
\end{figure}

\subsection{Layer-wise Representation}
Besides the vectorization quality and efficiency, the main objective of LIVE is to build a Layer-wise representation. Empirically, LIVE is able to explicitly vectorize each individual visual concept and explore the layer-wise representation for simple images like emojis and simple cliparts. We demonstrate the layer-wise representation ability of LIVE in Figure~\ref{fig:layerwise_easy}.
As shown in the figure, each component is clearly learned as a single \besizer path. Different from the vectorization methods that leverage segmentation pre-processing or use abundant paths, we can learn each component as an unbroken shape. In Figure~\ref{fig:4emoji}, we compare the vectorization results of LIVE and DiffVG on the Emoji benchmark. 

For complex images like photos and natural images, the topological clues are relatively hard to model. However, LIVE still exhibits a gratifying ability of the ``coarse-to-fine" learning style, as shown in Figure~\ref{fig:compare_diff_paths}. LIVE is more likely to achieve better reconstruction performance under the same number of paths. Moreover, we notice that LIVE models the local information much better than others, as shown in the red boxes. This may be explained by our progressive learning and the initialization method. In each step, LIVE encourages the new paths to fit the local details. 
While previous paths have successfully reconstructed the main context, the newly added path will only focus on initialized local regions with the enforcement of UDF loss. A comprehensive user study also demonstrated the superiority of LIVE (please refer to the supplementary).

\subsection{Interpolation}

Among existing vectorization methods, some VAE-based methods explored the application of interpolation~\cite{reddy2021im2vec,carlier2020deepsvg,lopes2019learned}. Even our LIVE is not based on the VAE model, we showcase that it is easy to achieve interpolation by integrating with vanilla raster image-based VAE model.

Before implementing the VAE interpolation, we first conduct an interesting interpolation experiment: given two semantically similar SVGs generated by LIVE, we directly interpolate the control points of each ordered path linearly. Normally, two SVGs are hard to be interpolated due to the disorder of both shapes and control points. In contrast, our LIVE will not suffer from this issue because of the ordered topology structures after optimization. Empirically, even with the simple interpolation of linear control points, LIVE still presents a reasonable result as shown in Figure~\ref{fig:interpolation}.

Next, we integrate our method with the VAE model. We train a simple VAE model on the MNIST dataset. Next, two random images are selected, we linearly interpolate the two latent vectors to obtain the interpolated images and use our LIVE to vectorize the resulted images sequence. To form a continuous sequence, we treat the previous result as the initialization of the next sample. Results in Figure~\ref{fig:interpolation} demonstrate that combing with a vanilla VAE model, our method works for interpolation as well. Given the efficient optimization method and great generalization ability, LIVE can be more practical to achieve the interpolation goal when combined with a powerful image generation model.

\subsection{Ablation study}
\paragraph{Circle Initialization.}
\begin{figure}
    \centering
    \includegraphics[width=0.9\linewidth]{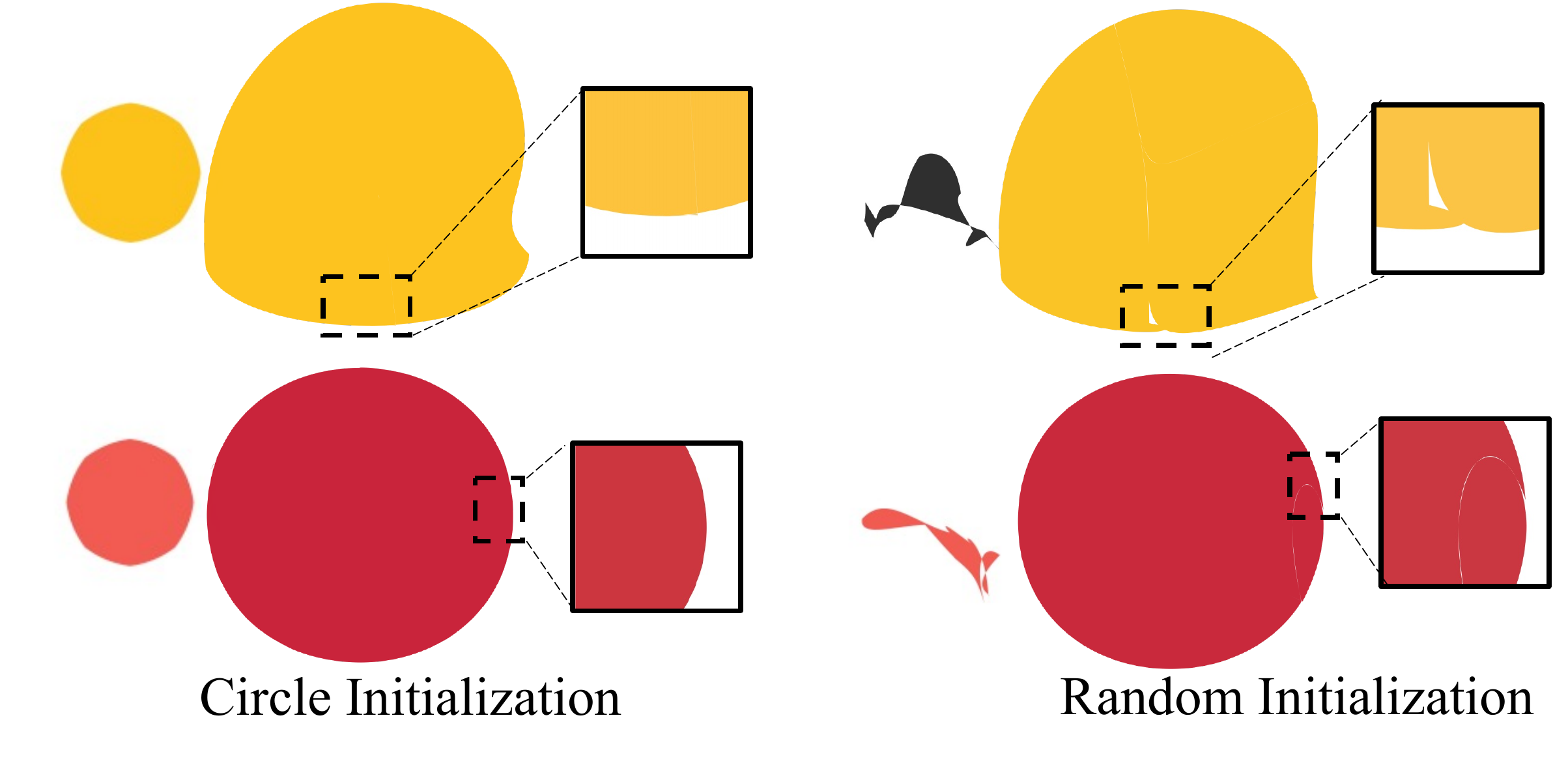}
    \caption{Examples of different initialization methods. For each triplet, we show the initialization (first column), output(second column), and the detail (third column). Zoom in to see better.}
    \label{fig:initialization}
    \vspace{-4mm}
\end{figure}
We first investigate the effectiveness of control point initialization.
Figure~\ref{fig:initialization} compares the circle initialization and random initialization.
Clearly, the circle initialization significantly reduces the artifacts compared with random initialization. Moreover, we notice that circle initialization is more likely to achieve better vectorization results, as shown in the first row. The reason is that with a circle initialization of control points, the close path is enforced to be convex, and gets a finer optimization result.

\begin{figure}
    \centering
    \includegraphics[width=0.9\linewidth]{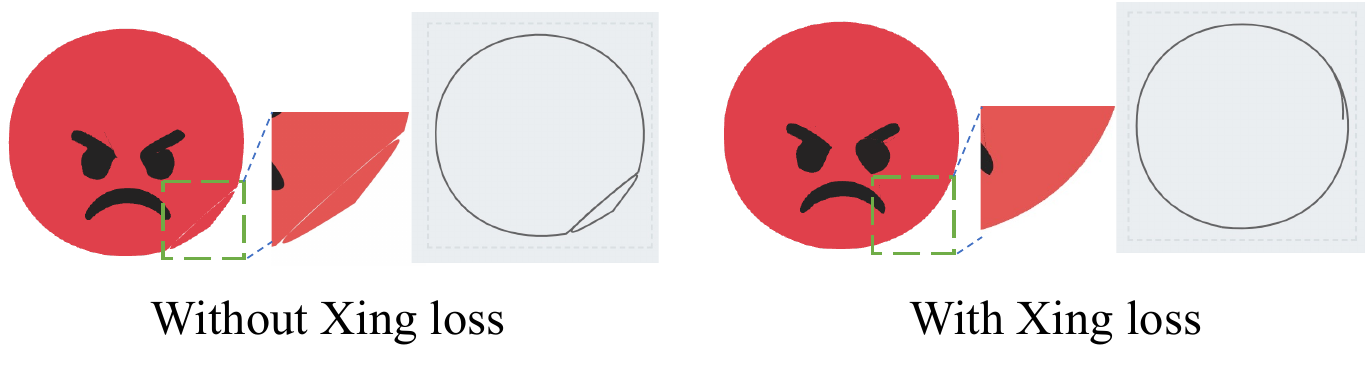}
    \caption{Illustration of the effectiveness of Xing loss. Each triplet shows the generated SVG, details, and the stroke of the face. By adding Xing loss, we greatly mitigate the self-interaction problem. Please zoom in to see the details. }
    \label{fig:xing}
      \vspace{-4mm}
\end{figure}

\paragraph{Xing Loss.}
To understand the effectiveness of the proposed Xing loss, we conduct an ablation study to investigate the impact of Xing loss through visualization in Figure~\ref{fig:xing}. 
With the help of Xing loss, we clearly mitigate the problem of self-interaction under the same optimization conditions. The circle shape tends to not intersect, given the constraints on the control points. It shows the proposed Xing loss is an intuitive, simple but effective objective function for mitigating the self-intersection issues. More results will be presented in the supplementary materials.

\subsection{Discussion} 
\paragraph{Limitations and Future Works.}
LIVE presents a layer-wise vectorization result, which can be used for further clipart creation or other applications. However, there are still some issues that we can discuss. First, the layer-wise operation is not efficient as the single-pass optimization. %
Some other methods also suffer from this issue~\cite{zou2021stylized}. An interesting research direction would be how we can combine the highly-efficient inference of deep models with the generalization ability of the optimization-based methods. Second, introducing gradient color and adaptively choosing the segment numbers and color type for each segment will be worth exploring. Third, for more complicated images like landscape or human photos, combining layer-wise vectorization with deep amodal segmentation in pixel space will be an interesting topic. We leave those for future works.  %

\paragraph{Potential Negative Impact.} Image to vector technologies can be misused by illegally converting and copying vector graph resources online, especially for easily reused and modified font or other images. To mitigate these issues, one can protect the copyright of the graph by using watermarks on raster images. Besides, though our paper achieves reasonable layer-wise modeling of the images, results converted from raster images can still be differentiated by checking whether each component is intact enough. Those actions will avoid the abuse of similar algorithms.

\section{Conclusion}\label{sec:conclusion}
In this work, we present Layer-wise Image VEctorization (LIVE), a framework to equip image vectorization with layer-wise representation. 
LIVE progressively infers the input raster image with the help of component-wise path initialization and new loss functions: an UDF loss for vectorization and a Xing loss to mitigate the self-interaction problem. With LIVE, we can explicitly vectorize individual components for a simple emoji or clipart, and investigate the ``coarse-to-fine" representation for complex natural images. To ease the evaluation of image vectorization, we also present two datasets, Emoji, and Pics. Besides image vectorization, LIVE can also be integrated with other methods to explore other applications like interpolation. 

{\small
\bibliographystyle{ieee_fullname}
\bibliography{egbib}
}

\appendix
\onecolumn
\section{User Study}
we conducted an user study\footnote{The details of the user study can be found at: \href{https://wj.qq.com/s2/9665341/19ed}{https://wj.qq.com/s2/9665341/19ed}.} to quantitatively compare our LIVE with DiffVG and Neural Painting. We randomly selected 21 images from emoji and pics datasets, and invited 20 users to select the method that has the best learning processing. The average scores of DiffVG, Neural Painting, and our LIVE are 14.3\%, 11.9\%, and 73.8\%, respectively. Results indicate that most people believe LIVE can achieve the best layer-wise representation.

\section{More Qualitative Comparisons} %
\begin{figure*}[!h]
    \centering
    \includegraphics[width=0.9\linewidth]{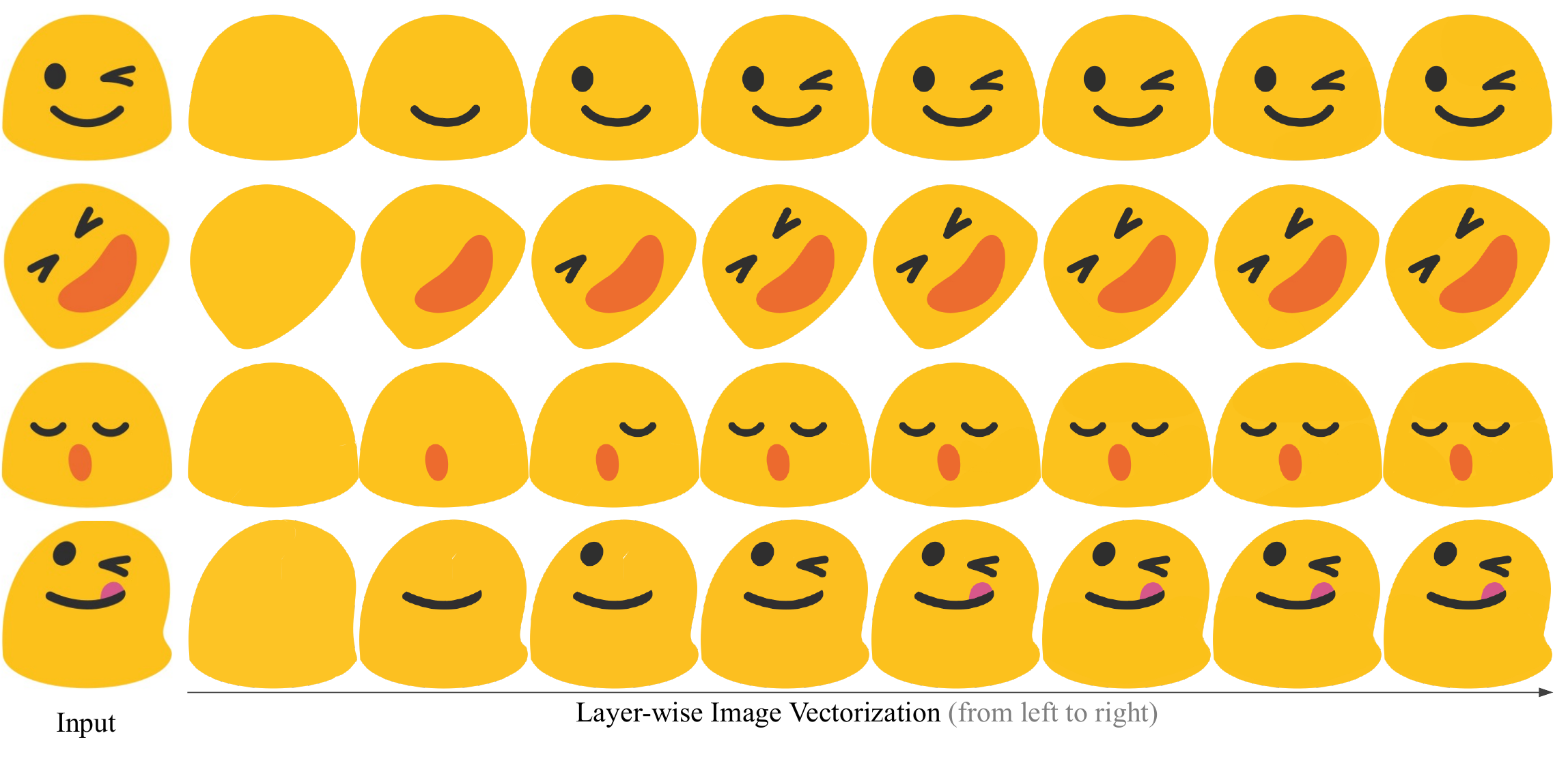}
    \caption{More examples of layer-wise representation. Given a simple image, our LIVE is able to learn each component in the image in a layer-wise fashion. Here we show the learning progress using 8 paths, where each output appends a new path to the previous result.}
    \label{fig:supp_more_emoji}
\end{figure*}

We also tested on more emoji images, as shown in Figure~\ref{fig:supp_more_emoji}. Without bells and whistles, LIVE explicitly disentangles the visual concepts, where each new path can fit a particular component in the input image. Adding more paths would not decrease the performance.

\section{Xing loss weights} %
\begin{figure}
    \centering
    \includegraphics[width=0.98\linewidth]{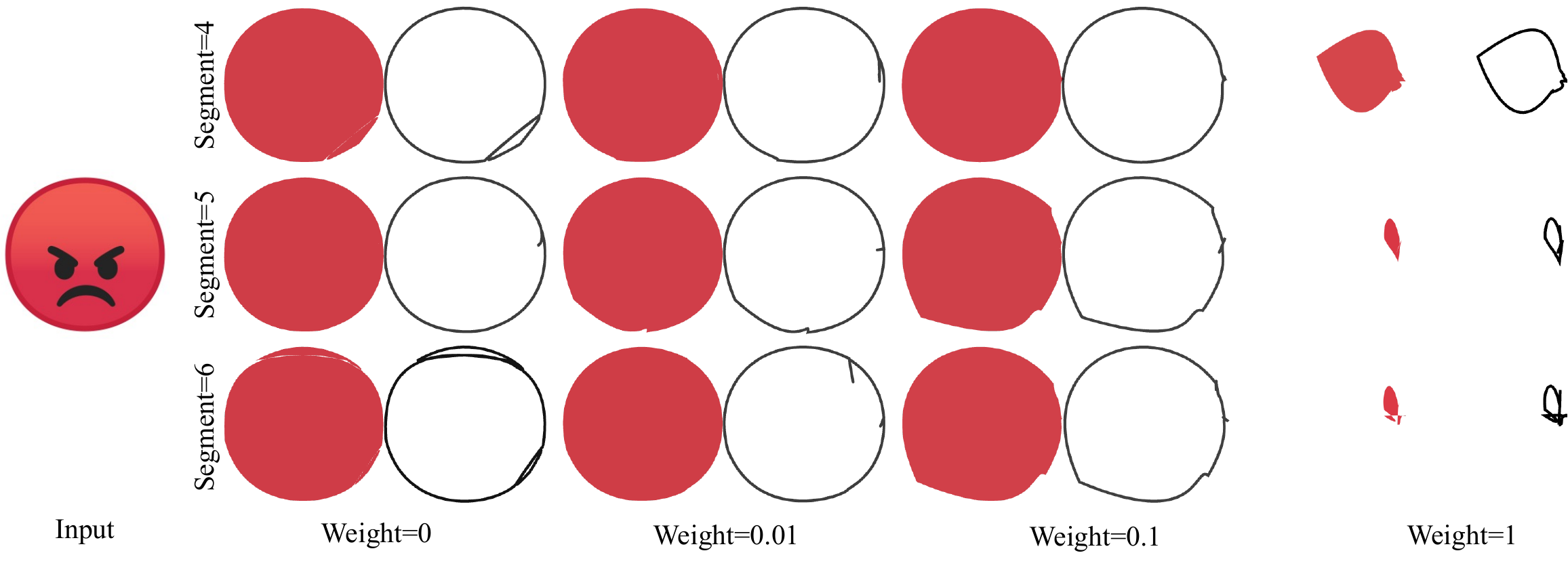}
    \caption{Impact of Xing loss weight. We showcase the results of first generated path with different segments number for illustration. In each pair, left is the generated SVG output and right is the corresponding \besizer curve.}
    \label{fig:supp_xing}
\end{figure}
We evaluate the impact of Xing loss in Figure~\ref{fig:supp_xing}. Generally, adding the Xing loss would greatly reduce the risk of self-interaction problems. We notice that a small weight of Xing loss can achieve the best result, while a larger weight (\textit{i.e.}, 1.0) always leads to the failure of optimization. Empirically, we set the Xing loss weight to 0.01 by default.

\section{Orderness of LIVE-generated SVGs}
An interesting property of LIVE-generated SVGs is the deterministic order of the optimized \besizer paths, due to the progressively learning pipeline and our component-wise initialization method. We demonstrate this property by linearly Interpolating two generated SVGs. We compare the results of DiffVG with rand seed, DiffVG with fixed seed, and our LIVE. Clearly, the interpolation results of DiffVG (with rand seed) are messed up because the path order is not deterministic. Even we fixed all randomness seeds, DiffVG still performs worse than our LIVE.
\begin{figure}
    \centering
    \includegraphics[width=0.7\linewidth]{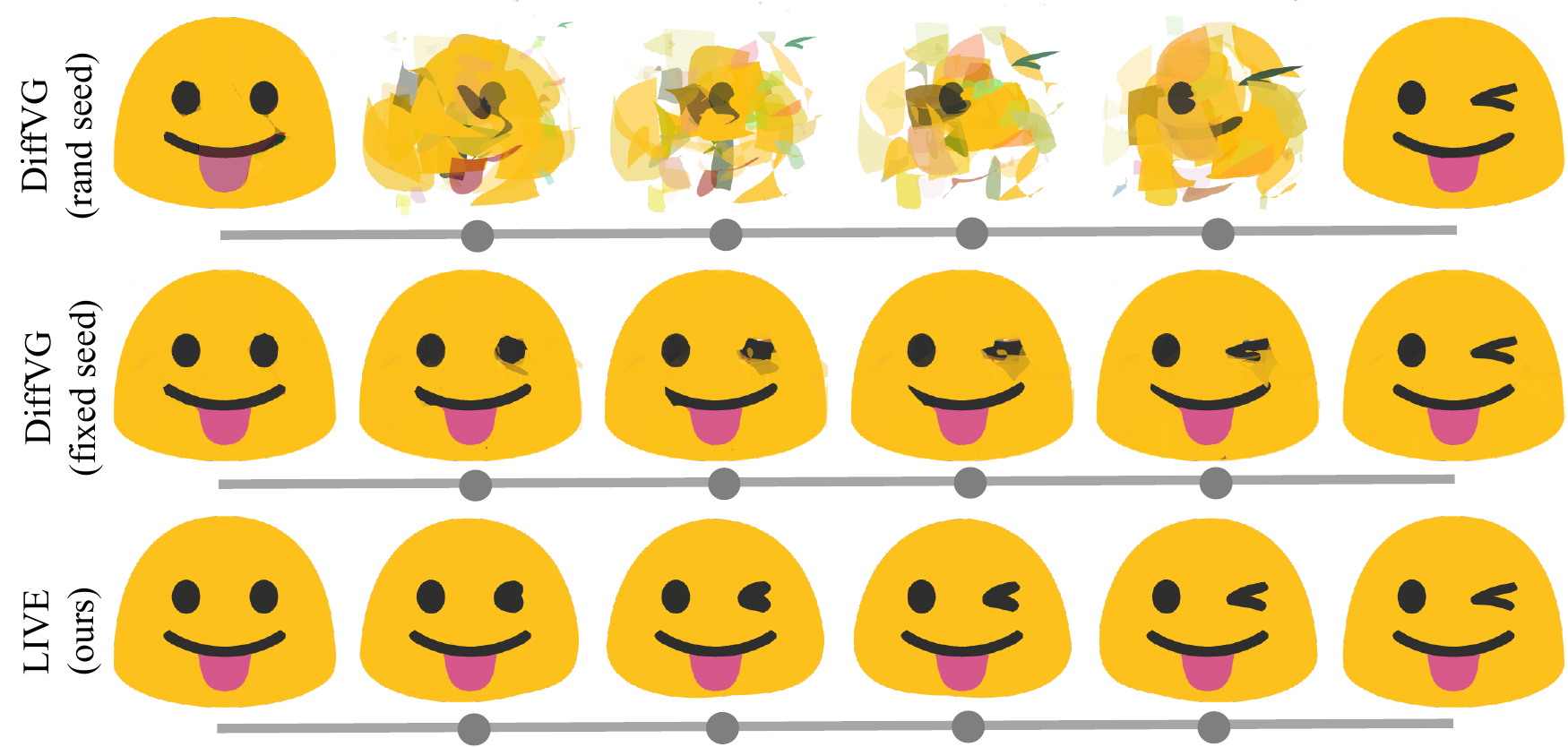}
    \caption{Linearly interpolating two SVGs generated by DiffVG (random seed), DiffVG (fixed seed), and our LIVE. }
    \label{fig:orderness}
\end{figure}

\section{More Interpolation Results}
\begin{figure}
    \centering
    \includegraphics[width=0.5\linewidth]{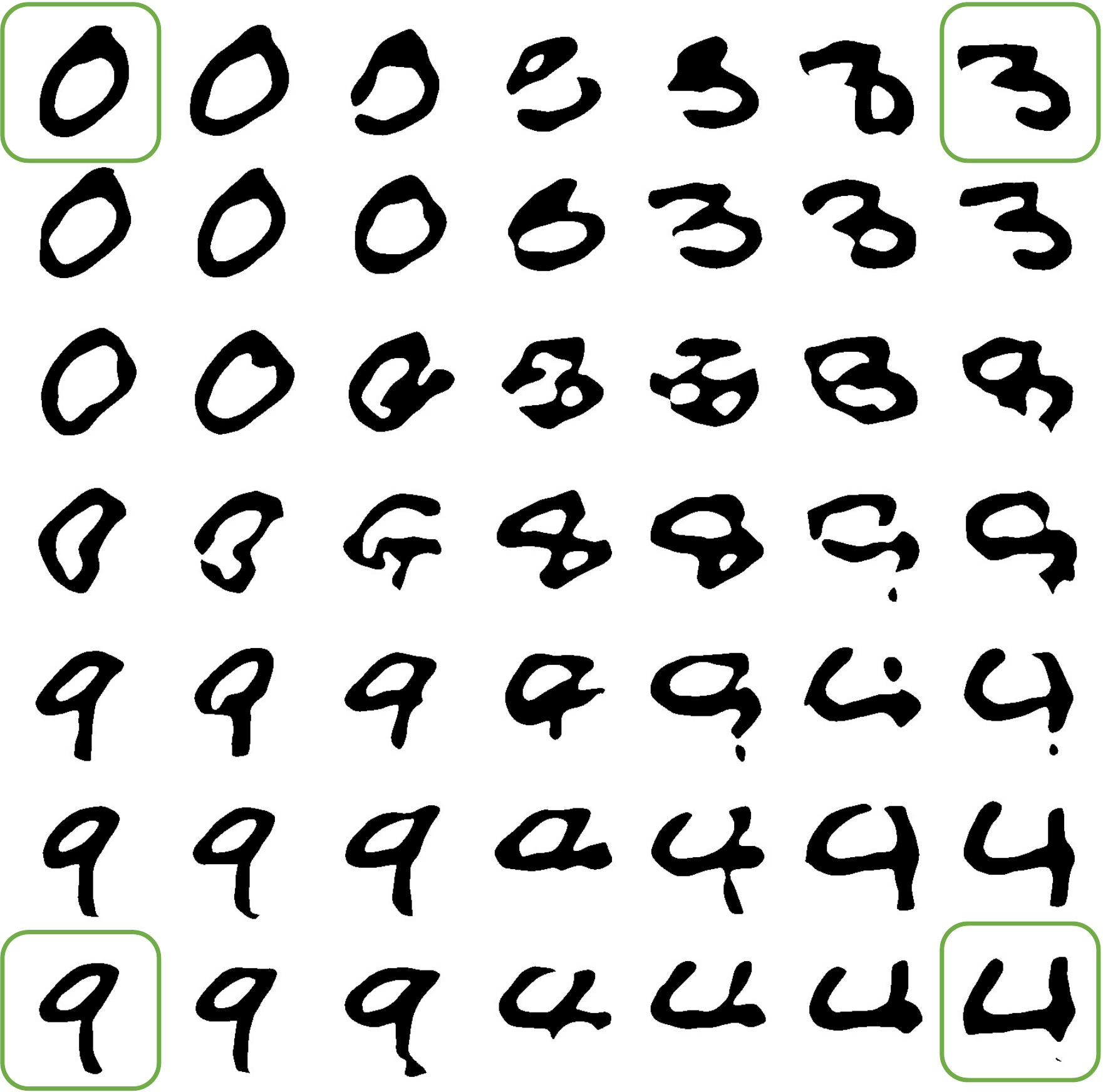}
    \caption{We plot the linear Interpolation results among four digits from MNIST dataset. We use the green boxes to emphasize the raster images. All the rest images are interpolated SVGs generated by the combination of VAE model and our LIVE method.}
    \label{fig:supp_mnist_matrix}
\end{figure}

We next present more interpolation results in Figure~\ref{fig:supp_mnist_matrix}. Rather than interpolating between two images, we further interpolate new images among four randomly selected images. Holistically, the results shown in Figure~\ref{fig:supp_mnist_matrix} indicate that combining a VAE model with our LIVE method can achieve similar results as other VAE-based vectorization methods.

\section{More Vectorization Quality} %
In this section, we present more examples that compare our LIVE with DiffVG and Neural Painting. 
The results are presented in the following figures, left (in the green box) is the input images, right is the output of different methods.
Empirically, under the same conditions (\textit{i.e.}, the number of paths/strokes), our LIVE can exhibit much better representation results, especially when the path number is small.  Please zoom in to see the details.

\begin{figure*}[!h]
\includegraphics[width=0.98\linewidth]{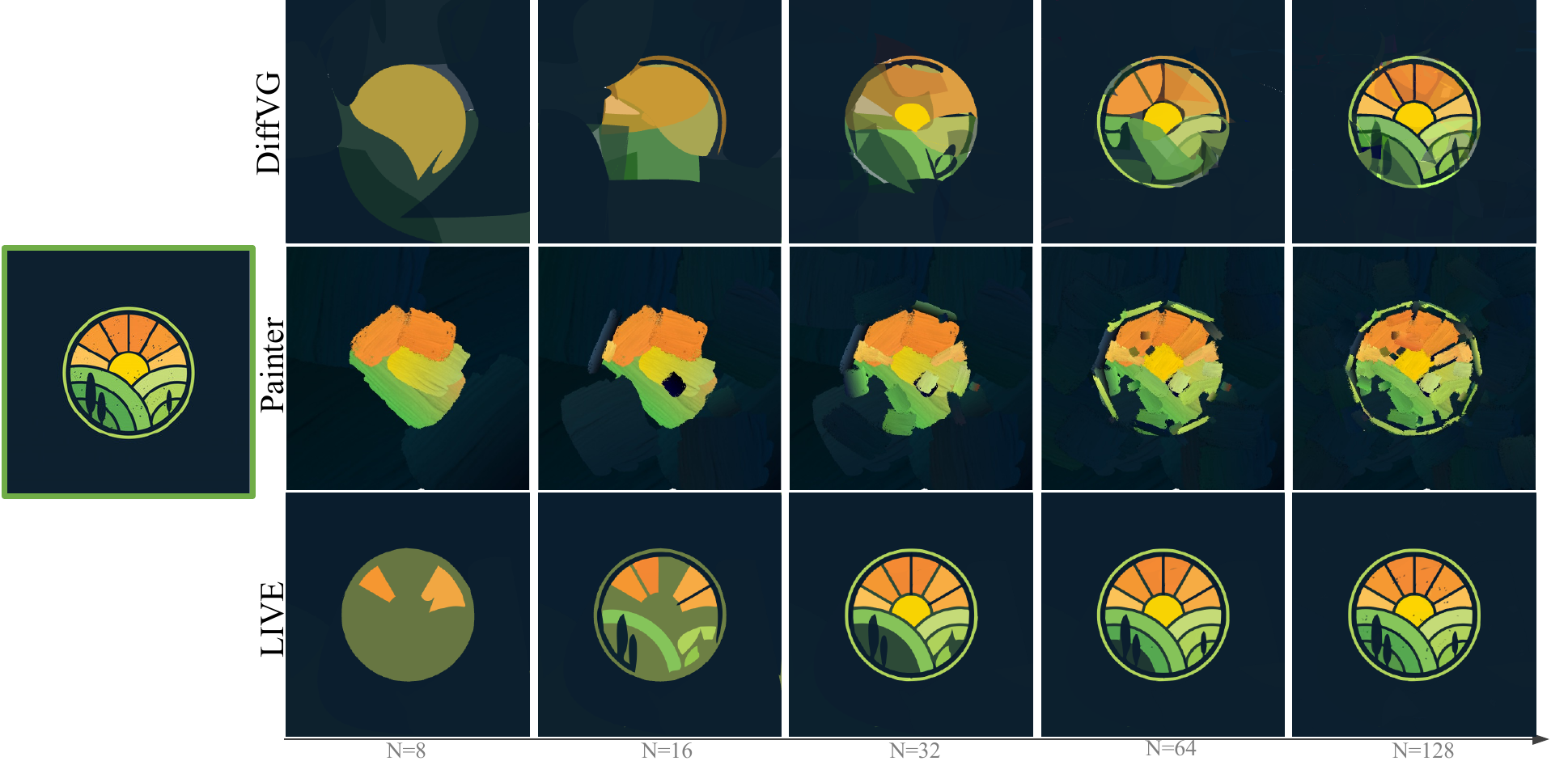}
\end{figure*}
\begin{figure*}[!h]
\includegraphics[width=0.98\linewidth]{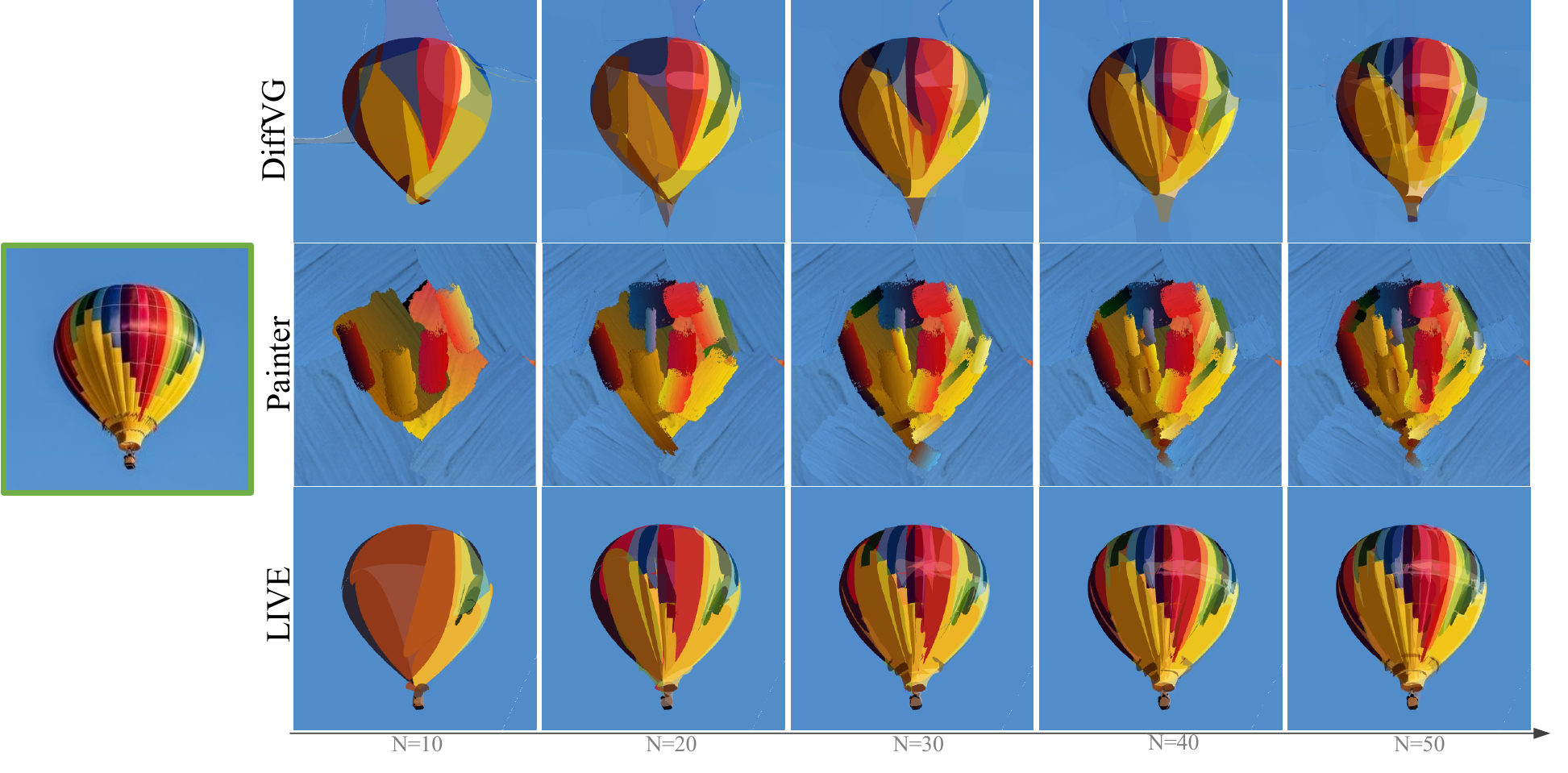}
\end{figure*}
\begin{figure*}[!h]
\includegraphics[width=0.98\linewidth]{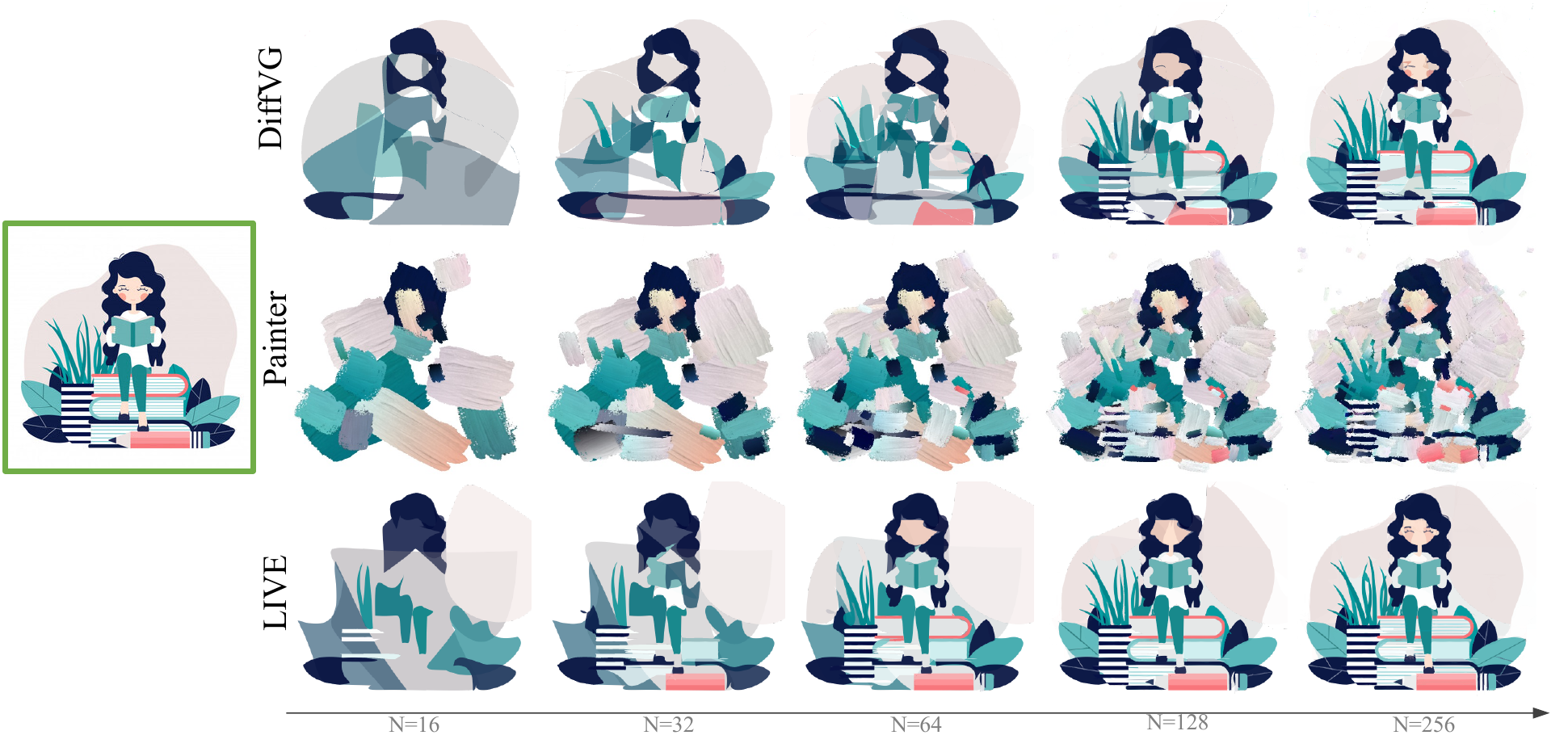}
\end{figure*}
\begin{figure*}[!h]
\includegraphics[width=0.98\linewidth]{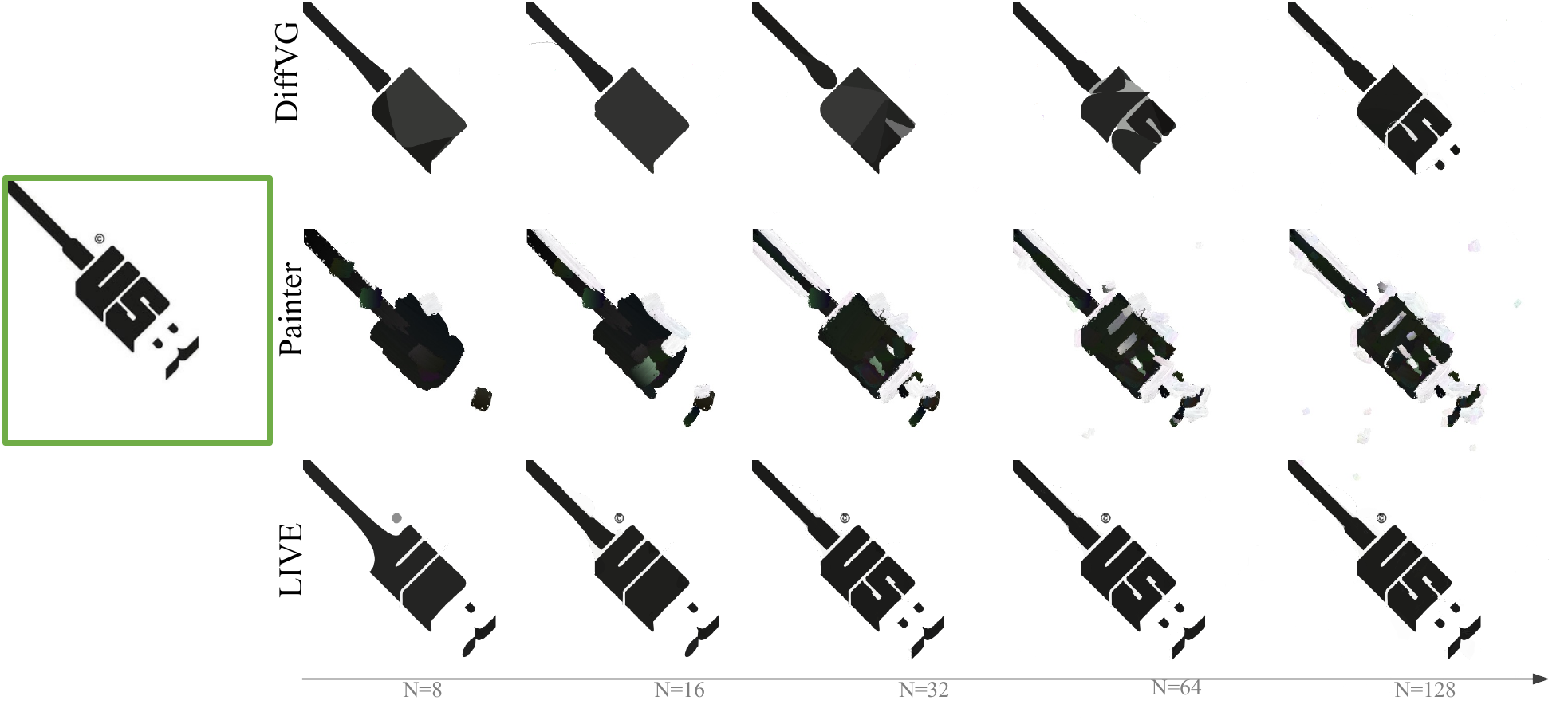}
\end{figure*}
\begin{figure*}[!h]
\includegraphics[width=0.98\linewidth]{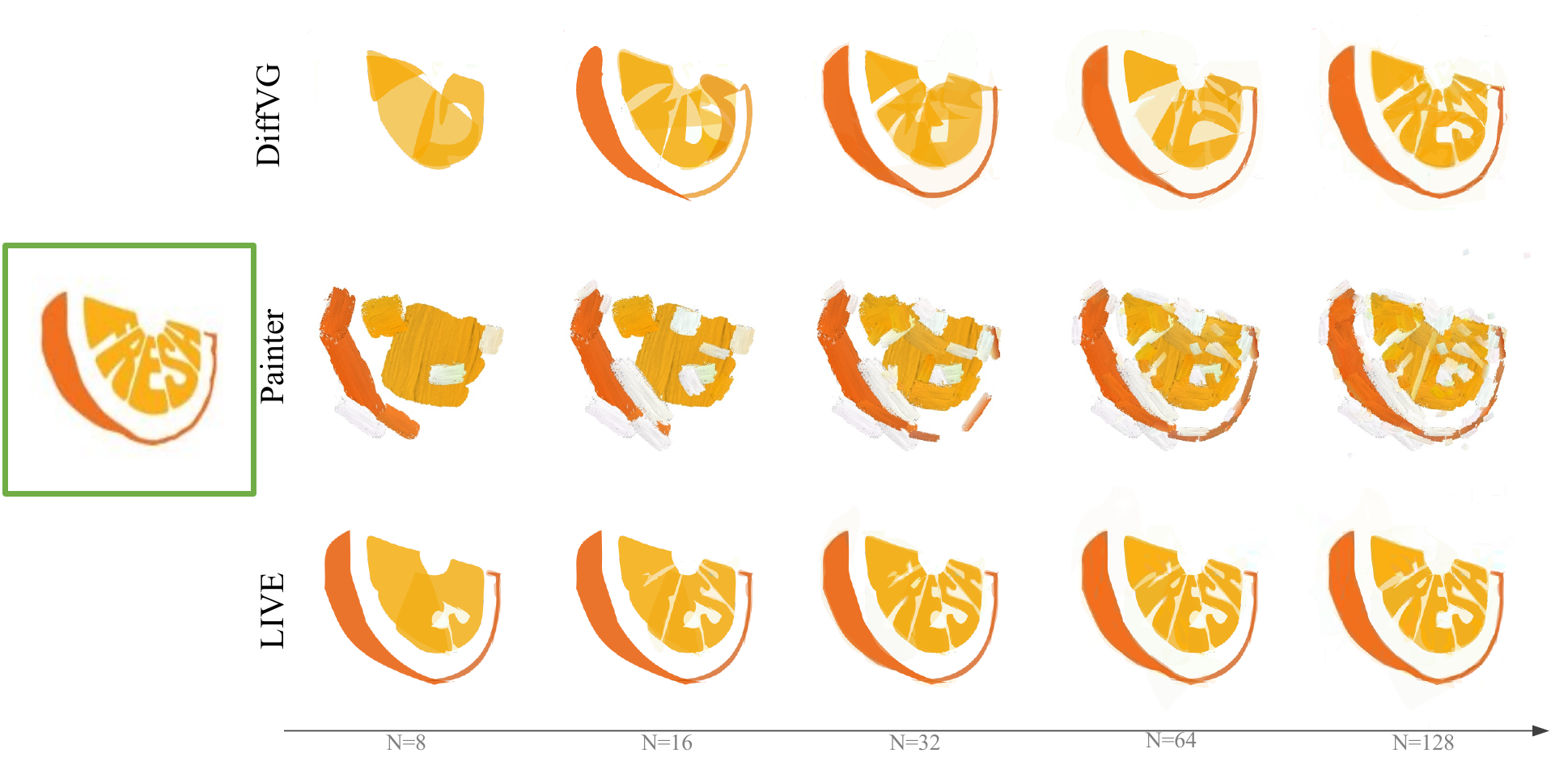}
\end{figure*}
\begin{figure*}[!h]
\includegraphics[width=0.98\linewidth]{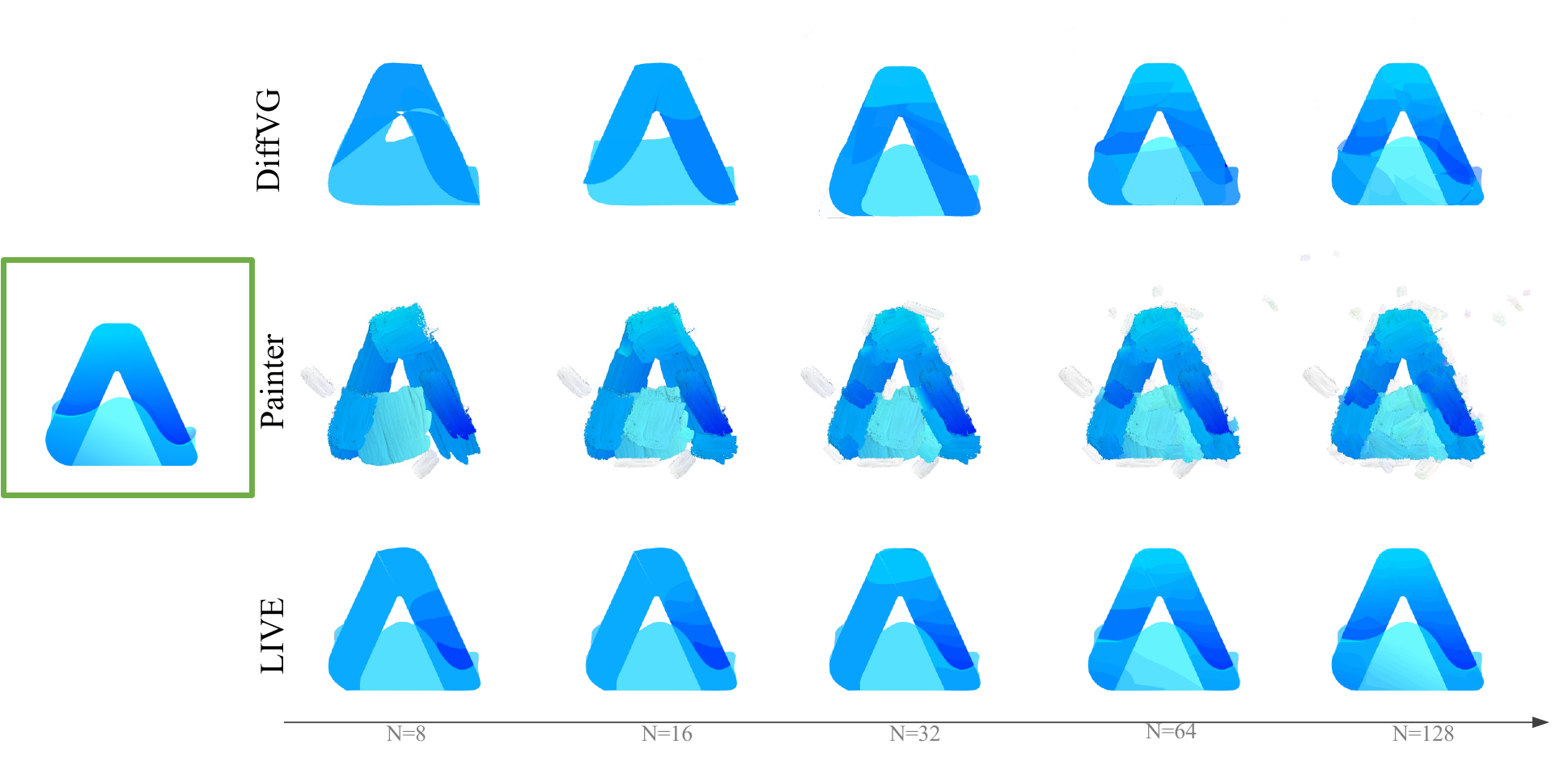}
\end{figure*}
\begin{figure*}[!h]
\includegraphics[width=0.98\linewidth]{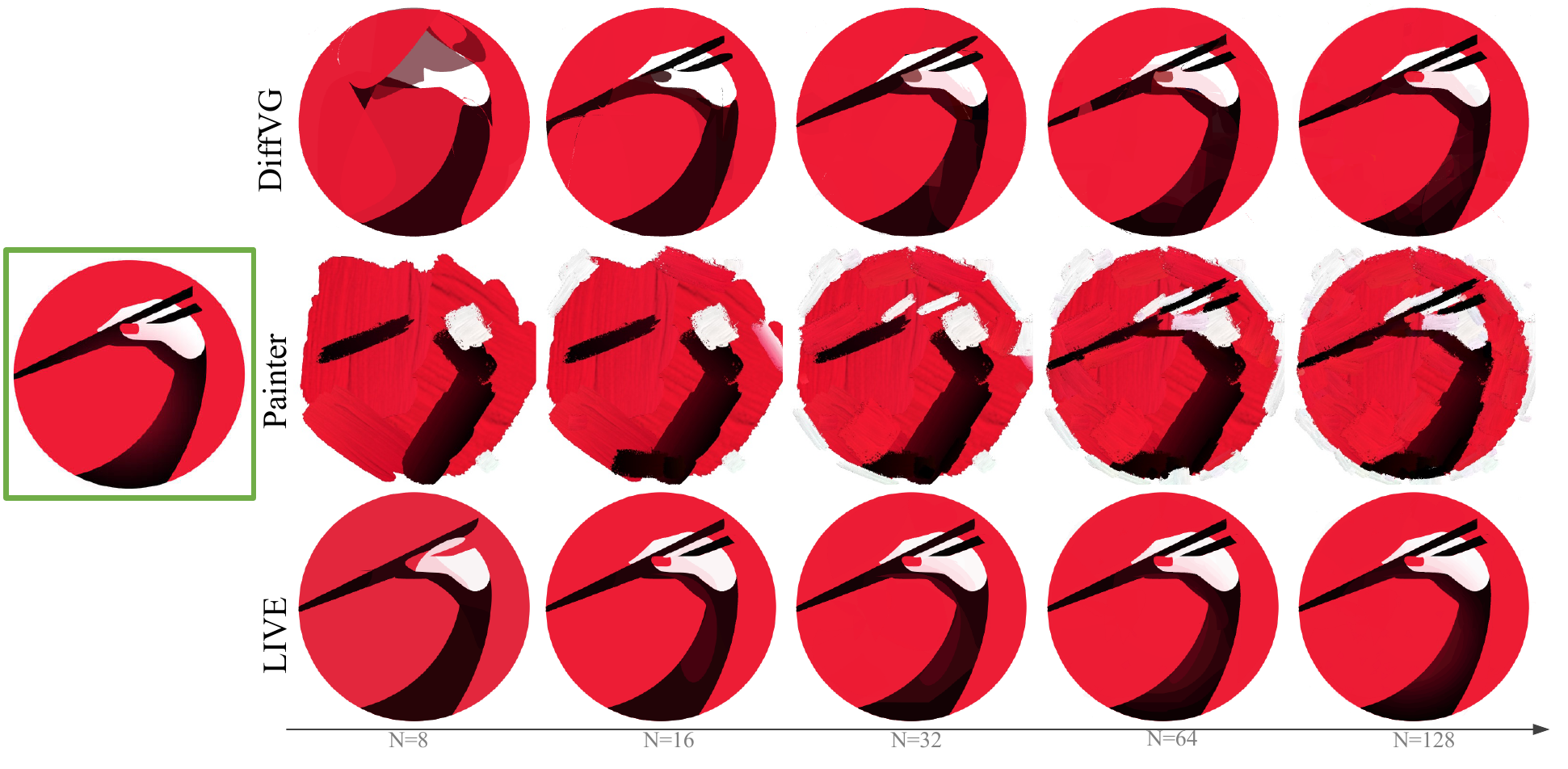}
\end{figure*}
\begin{figure*}[!h]
\includegraphics[width=0.98\linewidth]{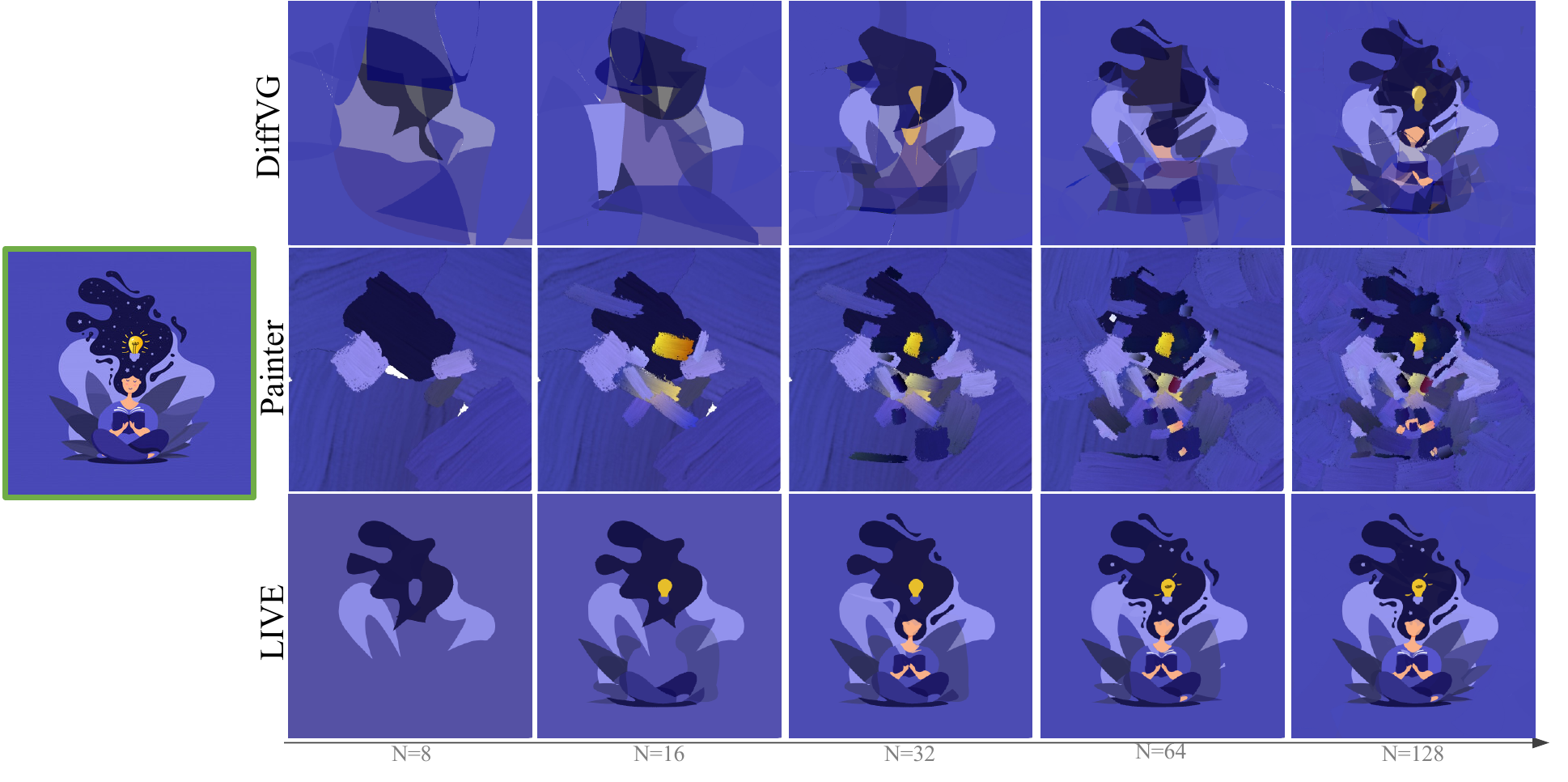}
\end{figure*}
\begin{figure*}[!h]
\includegraphics[width=0.98\linewidth]{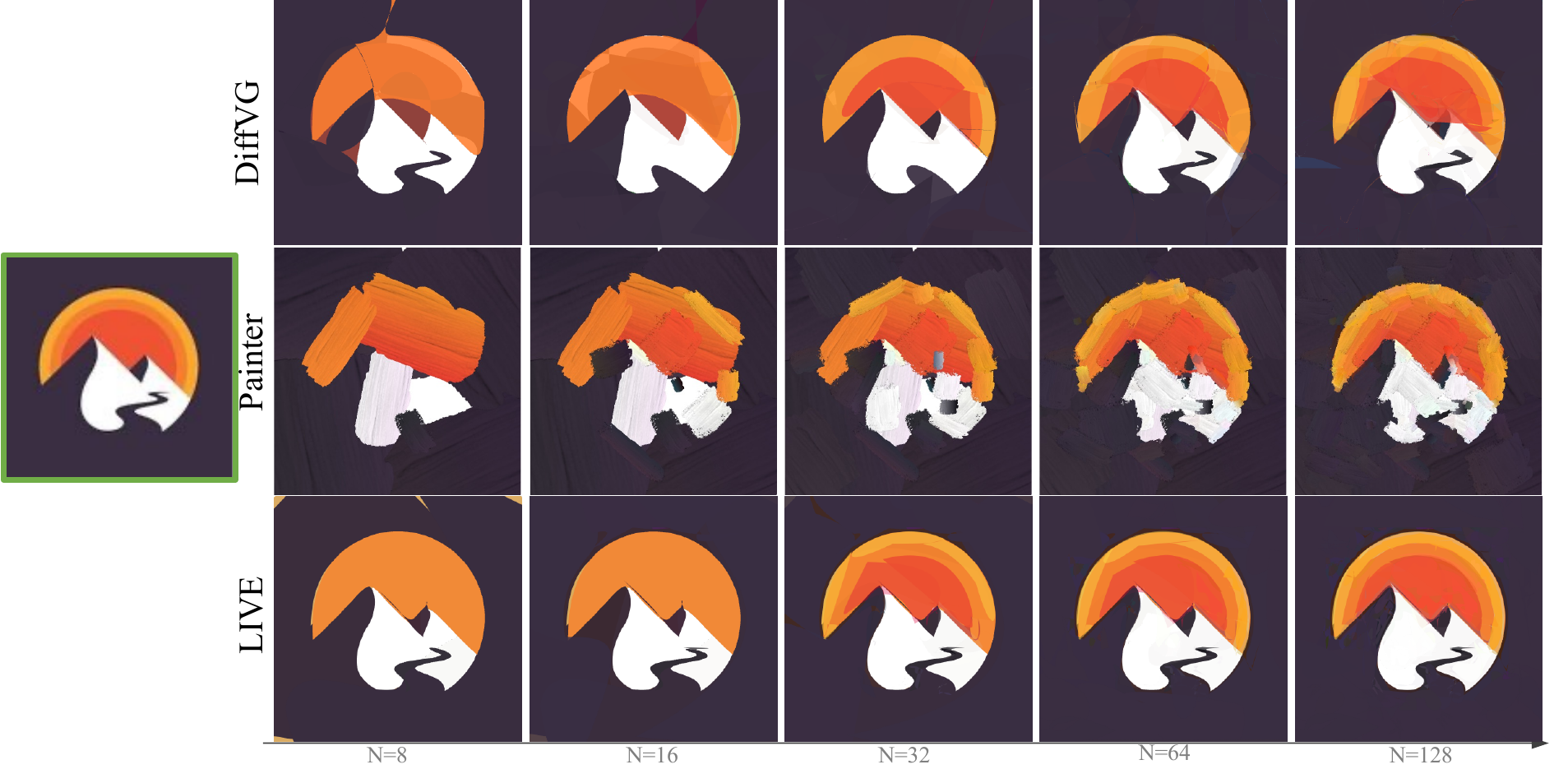}
\end{figure*}
\begin{figure*}[!h]
\includegraphics[width=0.98\linewidth]{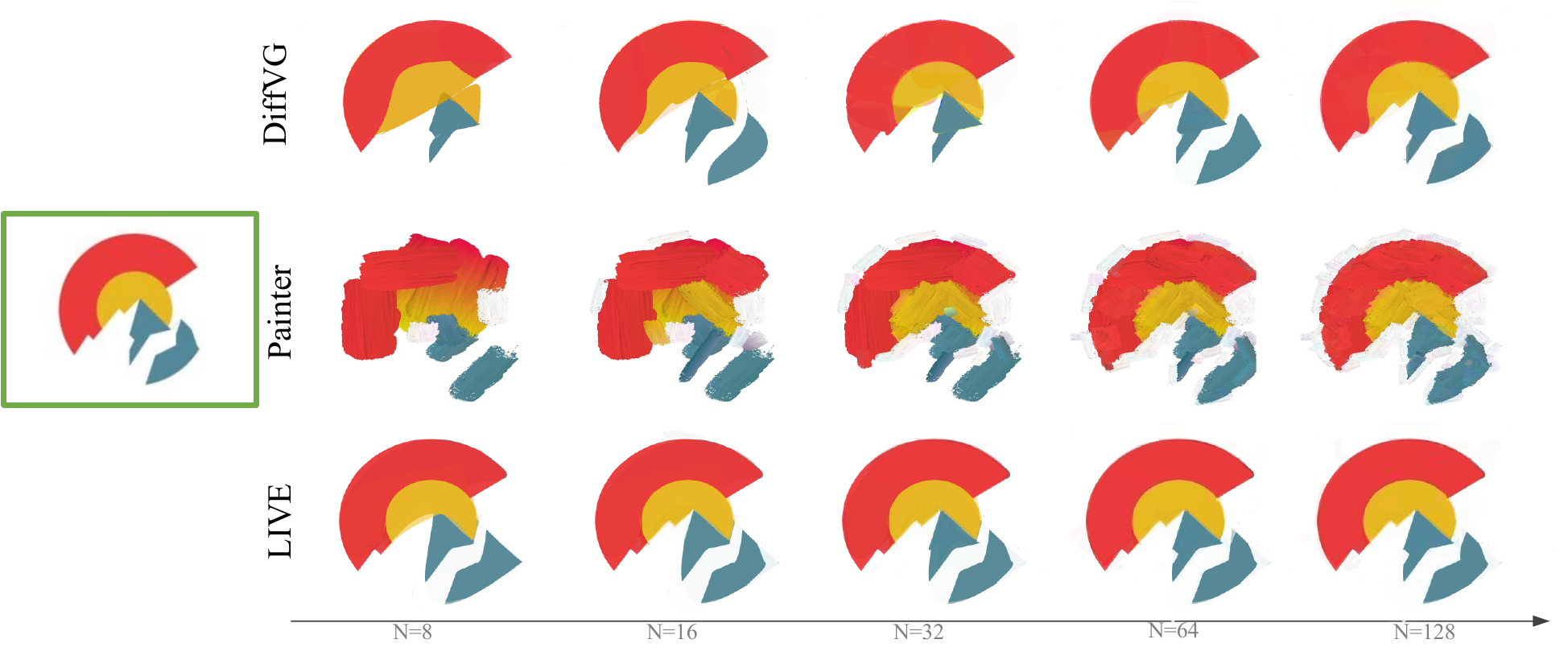}
\end{figure*}
\begin{figure*}[!h]
\includegraphics[width=0.98\linewidth]{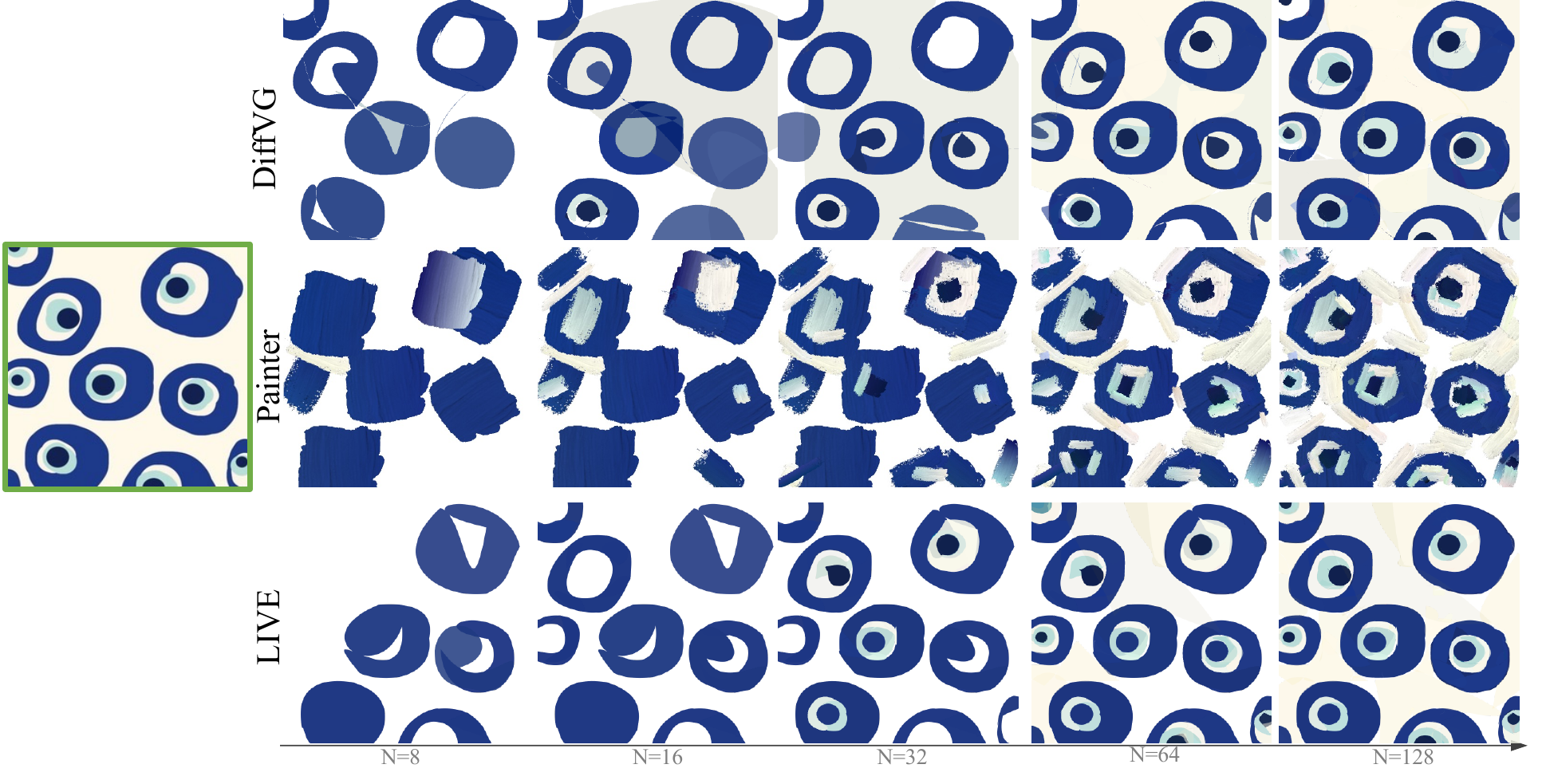}
\end{figure*}
\begin{figure*}[!h]
\includegraphics[width=0.98\linewidth]{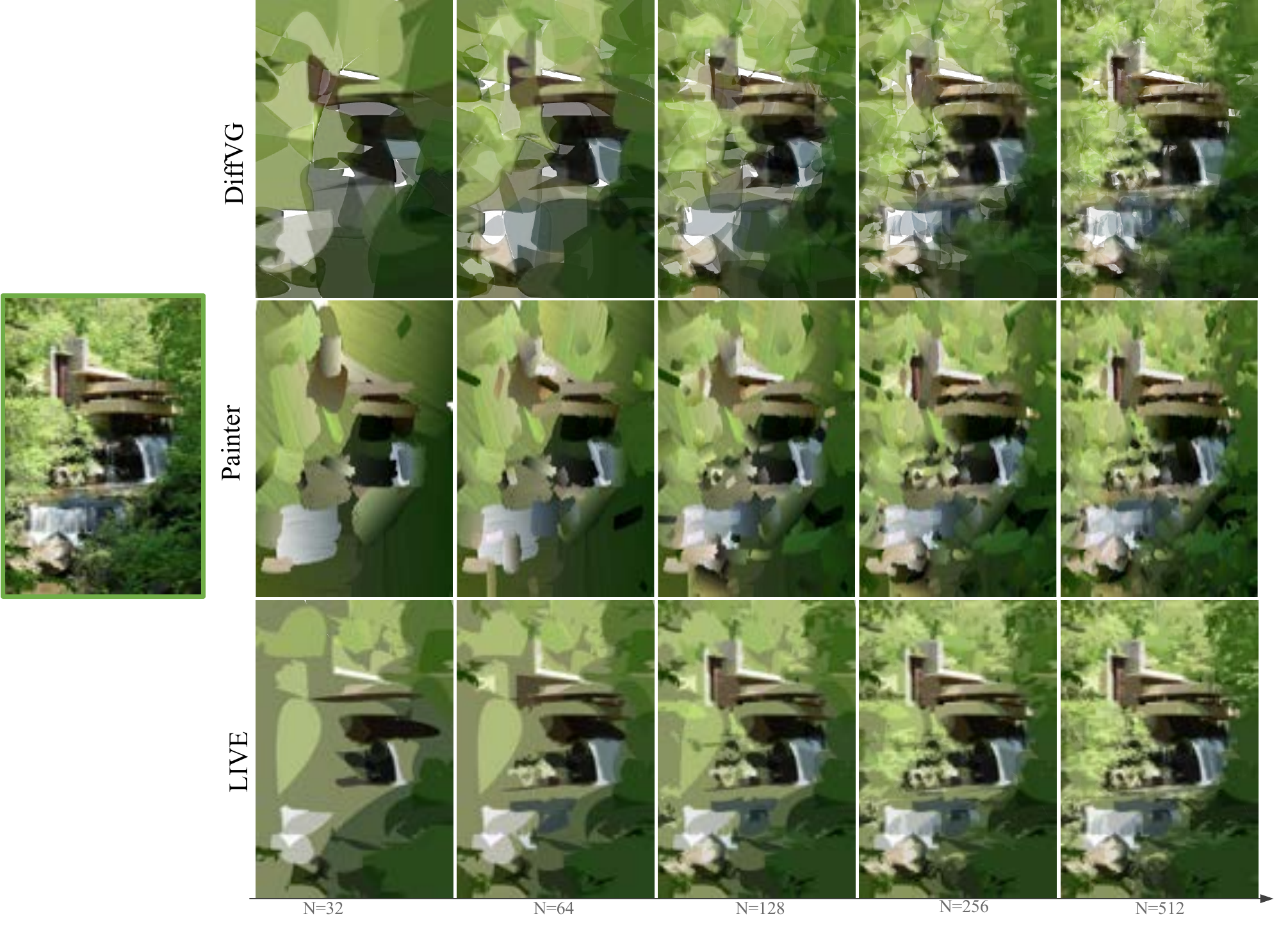}
\end{figure*}

\end{document}